\documentclass[sigconf, screen]{acmart}
\AtBeginDocument{%
  }

\setcopyright{acmlicensed}
\copyrightyear{2018}
\acmYear{2018}
\acmDOI{XXXXXXX.XXXXXXX}
\acmConference[Conference acronym 'XX]{Make sure to enter the correct
  conference title from your rights confirmation email}{XX XX--XX,
  XX}{XX, XX}
\acmISBN{978-1-4503-XXXX-X/2018/06}

\acmSubmissionID{4282}

\usepackage[table]{xcolor}
\usepackage{booktabs}
\usepackage{multirow}
\usepackage{enumitem}
\usepackage{makecell}
\newcommand{\gain}[1]{\textcolor{green!60!black}{(+#1)}}

\begin{document}

\title{RefereeBench: Are Video MLLMs Ready to be Multi-Sport Referees?}

\author{Yichen Xu}
\email{xu_yichen@ruc.edu.cn}
\affiliation{%
  \institution{Renmin University of China}
  \city{Beijing}
  \country{China}
}

\author{Yuanhang Liu}
\email{2023202275@ruc.edu.cn}
\affiliation{%
  \institution{Renmin University of China}
  \city{Beijing}
  \country{China}
}

\author{Chuhan Wang}
\email{wangchuhan51@stu.scu.edu.cn}
\affiliation{%
  \institution{Sichuan University}
  \city{Chengdu}
  \country{China}
}

\author{Zihan Zhao}
\email{zh1579@ruc.edu.cn}
\affiliation{%
  \institution{Renmin University of China}
  \city{Beijing}
  \country{China}
}

\author{Jinghan Luo}
\email{2024201381@ruc.edu.cn}
\affiliation{%
  \institution{Renmin University of China}
  \city{Beijing}
  \country{China}
}

\author{Jianzhe Ma}
\email{majianzhe@ruc.edu.cn}
\affiliation{%
  \institution{Renmin University of China}
  \city{Beijing}
  \country{China}
}

\author{Wenxuan Wang}
\authornote{Qin Jin and Wenxuan Wang are the corresponding authors.}
\email{wangwenxuan@ruc.edu.cn}
\affiliation{%
  \institution{Renmin University of China}
  \city{Beijing}
  \country{China}
}

\author{Qin Jin}
\authornotemark[1]
\email{qjin@ruc.edu.cn}
\affiliation{%
  \institution{Renmin University of China}
  \city{Beijing}
  \country{China}
}

\renewcommand{\shortauthors}{Yichen Xu, Yuanhang Liu, Chuhan Wang et al.}
\renewcommand\footnotetextcopyrightpermission[1]{} 
\settopmatter{printacmref=false} 

\begin{abstract}
    While Multimodal Large Language Models (MLLMs) excel at generic video understanding, their ability to support specialized, rule-grounded decision-making remains insufficiently explored. In this paper, we introduce \textbf{RefereeBench}, the first large-scale benchmark for evaluating MLLMs as automatic sports referees. Spanning \textbf{11 sports} with 925 curated videos and 6,475 QA pairs, RefereeBench evaluates five core officiating abilities: foul existence, foul and penalty classification, foul and penalty reasoning, entity perception, and temporal grounding. The benchmark is \textbf{fully human-annotated} to ensure high-quality annotations grounded in authentic officiating logic and multimodal evidence. Extensive evaluations of state-of-the-art MLLMs show that even the strongest models, such as Doubao-Seed-1.8 and Gemini-3-Pro, achieve only around \textbf{60\%} accuracy, while the strongest open-source model, Qwen3-VL, reaches only \textbf{47\%}. These results indicate that current models remain far from being reliable sports referees. Further analysis shows that while models can often identify incidents and involved entities, they struggle with rule application and temporal grounding, and frequently over-call fouls on normal clips. Our benchmark highlights the need for future MLLMs that better integrate domain knowledge and multimodal understanding, advancing trustworthy AI-assisted officiating and broader multimodal decision-making.
\end{abstract}

\ccsdesc[500]{Computing methodologies~Natural language processing}
\ccsdesc[500]{Computing methodologies~Computer vision}

\keywords{Automatic Sports Refereeing, Multimodal Large Language Models, Sports Understanding}



\maketitle

\section{Introduction}

\begin{figure*}[htbp]
  \includegraphics[width=0.95\textwidth]{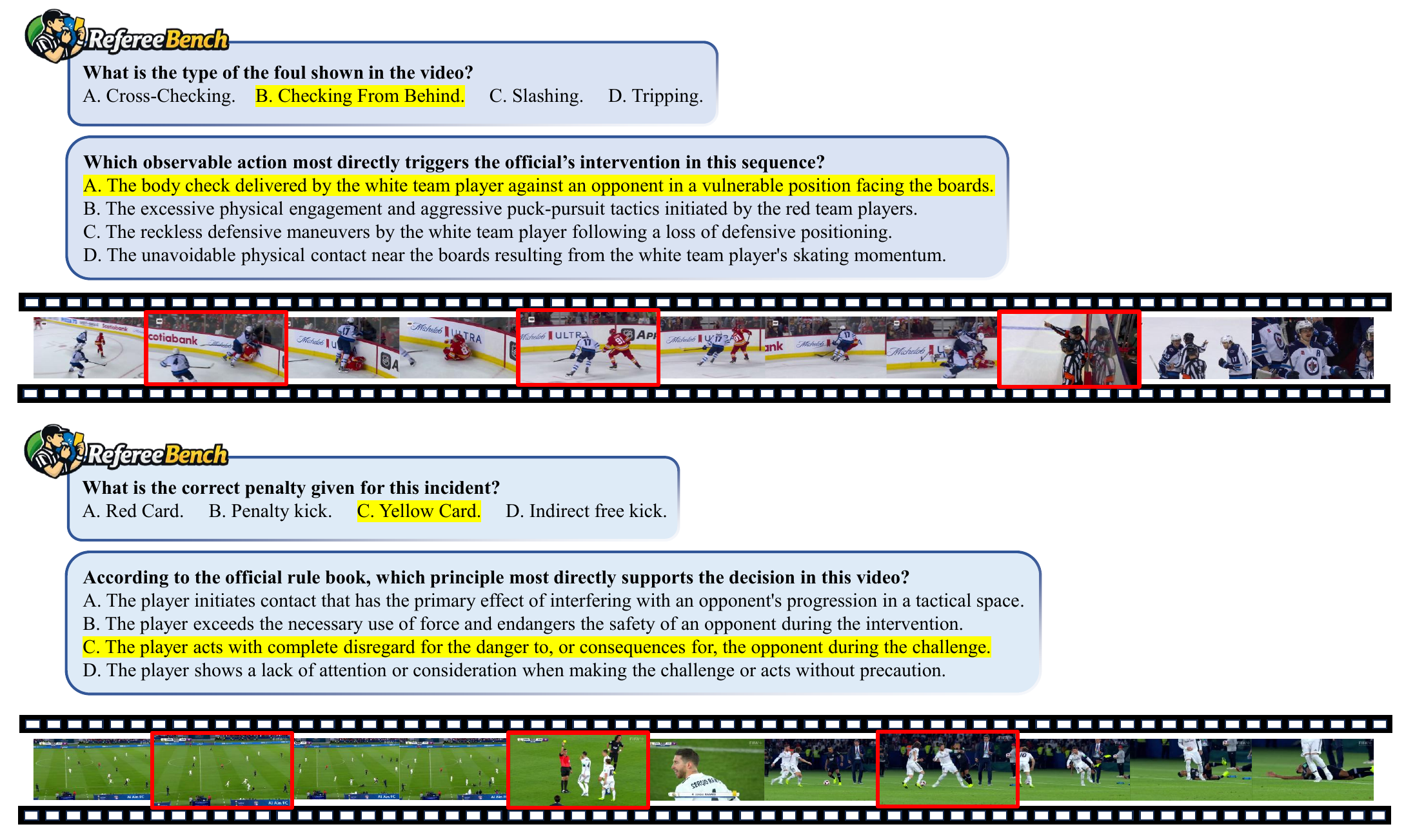}
  \caption{We introduce RefereeBench to provide high-quality assessment of MLLMs in automatic sports refereeing, with all videos and annotations manually curated by certified referees. These two highlighted examples of RefereeBench (ground-truth answers in \textcolor{yellow!90!black}{yellow}) that require models to perform fine-grained perception, temporal grounding, rule knowledge application and decision-making. RefereeBench is designed to pose significant challenges, effectively evaluating the multi-sport officiating and reasoning capabilities of MLLMs.}
  \label{fig:s}
\end{figure*}

Sports have become an important frontier for multimodal AI, where visual understanding, temporal reasoning, and decision support increasingly meet real-world needs\cite{xusurvey, xia2024language, zhao2025survey, soccernet, soccernet-v2, cui2023sportsmot, sun2020tracknetv2, ingwersen2023sportspose, falaleev2024enhancing, wang2024tacticai, soccernet-caption, qi2023goal, you2025timesoccer, gao2025fsbench, xu2022finediving, tian2025sportsgpt, rao2025multi}. Among sports applications, officiating is particularly consequential, since refereeing decisions directly affect match fairness, athlete safety, and public trust\cite{xusurvey}. More importantly, intelligent refereeing assistance is valuable not only in professional tournaments, but also in amateur and community games, where experienced referees and replay infrastructure are often limited\cite{xusurvey, xia2024language}. This makes automatic sports refereeing both scientifically challenging and practically meaningful.

However, automatic sports refereeing is substantially more difficult than generic video understanding. A competent system must do more than recognize visible actions: it must determine whether a violation occurs, identify the involved entities, localize the decisive moment, distinguish visually similar but legally different events, and relate the observed evidence to sport-specific rules and penalty standards\cite{xusurvey}. In short, the task requires not only perception, but also temporal grounding and rule-grounded reasoning.

The rapid progress of Multimodal Large Language Models (MLLMs) brings automatic sports refereeing into focus as an important testbed. Although recent models perform strongly in multimodal reasoning, long-video understanding, and fine-grained event recognition\cite{li2024mvbench, fu2025video, liu2025videoreasonbench, wu2024longvideobench, xun2025rtv, feng2025breaking}, these capabilities do not necessarily translate into reliable rule-grounded decision-making. Prior benchmarks show that current models still struggle with consistent decisions in complex video scenarios, especially when judgments require structured reasoning and domain-specific knowledge\cite{li2024mvbench, fu2025video, song2025video, xia2024sportu}. Sports refereeing therefore offers a natural setting to evaluate the gap between general multimodal competence and real-world decision-making.

Despite this promise, existing datasets and benchmarks still do not adequately evaluate automatic sports refereeing as a general problem. Prior sports benchmarks mainly focus on action and event spotting\cite{soccernet, soccernet-v2, shao2020finegym}, commentary\cite{soccernet-caption, ge2024scbench}, action quality assessment\cite{xu2022finediving, gao2025fsbench}, or general sports reasoning\cite{xia2024sportu, xia2024sportqa}, while refereeing-oriented resources are typically limited to a single sport, especially soccer, or do not cover the full officiating scope\cite{soccernet, held2023vars, held2024xvars}. As a result, it remains unclear whether current MLLMs can serve as reliable refereeing assistants across multiple sports with different rule systems, event structures, and sanction logics.

To address this gap, we introduce \textbf{RefereeBench}, the \textbf{first} large-scale benchmark dedicated to automatic sports refereeing across multiple sports. RefereeBench contains 925 curated real-match videos and 6,475 expert-authored multiple-choice QA pairs spanning 11 sports. It evaluates \textbf{seven officiating-oriented question types}, covering foul existence, foul type recognition, foul reasoning, penalty type recognition, penalty reasoning, entity perception, and incident temporal grounding. Moreover, the benchmark preserves realistic multimodal signals, including audio and replay views, to better reflect real-world officiating scenarios. 

Built on RefereeBench, we conduct a comprehensive evaluation of a diverse set of proprietary and open-source MLLMs. Our study covers both closed-source systems, including GPT\cite{openai_gpt5_2025, openai_gpt4o_2024}, Gemini\cite{gemini3_google_2025}, Claude\cite{claude45_anthropic_2025}, and Doubao-Seed\cite{bytedance2025seed18}, and open-source video models such as Qwen3-VL\cite{xu2025qwen3} and VideoLLaMA3\cite{zhang2025videollama}. This evaluation aims to address a central question: \textbf{to what extent are current MLLMs capable of reliable, multi-sport officiating?} The results show that even the strongest models remain far from dependable referees, with only Doubao-Seed-1.8 and Gemini-3 reaching the performance of around 60\%. Open-source models lag behind, with Qwen3-VL achieving only 47\%. Although they achieve relatively stronger performance on foul existence and entity perception, they still struggle substantially on reasoning and temporal grounding, suggesting that recognizing an incident is much easier than applying the correct sports rules and identifying decisive evidence in time. We further analyze why current models exhibit such performance patterns and probe what contributes most to their automatic refereeing ability. We find that multimodal evidence, especially audio, plays a critical role in improving performance, while models possess only partial refereeing knowledge and remain sensitive to misleading or suggestive inputs. In addition, they show a clear tendency to over-call fouls, highlighting limitations in robust and unbiased decision-making.

In summary, this paper makes three main contributions:
\begin{itemize}
    \item We introduce \textbf{RefereeBench}, the first large-scale benchmark for automatic sports refereeing across multiple sports, enabling systematic evaluation beyond single-sport settings.
    \item We construct a high-quality, \textbf{expert-annotated multimodal dataset} spanning 11 sports and seven officiating-oriented question types, capturing realistic refereeing evidence such as audio, replay views, and temporal grounding.
    \item We conduct a \textbf{comprehensive evaluation and analysis} of modern MLLMs on rule-grounded sports decision-making, revealing key limitations in reasoning, grounding, and robustness, and highlighting the gap between perception and reliable refereeing.
\end{itemize}
\section{Related Work}

\subsection{Sports Understanding Datasets}
Early sports video analysis primarily targeted foundational perception. Initial benchmarks focused on action spotting and fine-grained recognition\cite{soccernet, soccernet-v2, shao2020finegym, liu2025f, he2025finebadminton} to localize and classify player or ball movements. Subsequent research expanded to capture motion details via multi-object tracking\cite{cui2023sportsmot, huang2019tracknet, sun2020tracknetv2} and 3D pose estimation\cite{ingwersen2023sportspose}. Building on these perceptual foundations, action quality assessment datasets\cite{xu2022finediving, gao2025fsbench} emerged to evaluate the execution of fine-grained motions, acting as an early bridge toward objective performance evaluation. As multimodal architectures evolved, the paradigm shifted toward comprehensive semantic understanding. A significant line of work focuses on generating rich narratives, including automated commentary\cite{soccernet-caption, ge2024scbench}, sports news\cite{wang2021sportssum2}, and highlight extraction\cite{diaz2025soccerhigh}. Concurrently, complex domain question-answering benchmarks such as Sports-QA\cite{li2024sports}, SportQA\cite{xia2024sportqa} and SportU\cite{xia2024sportu} were introduced to systematically test models' holistic comprehension of basic rules, game scenarios, and tactical intents. Most recently, researchers have begun exploring decision-level analytics requiring higher-order cognitive reasoning\cite{zhao2025survey}, such as tactical analysis\cite{wang2024tacticai}, game prediction\cite{honda2022pass}, and early foul detection\cite{soccernet-v2, held2023vars}. Although modern MLLMs possess the advanced reasoning required for practical, decision-centric applications, current literature lacks benchmarks evaluating strict, cross-sport refereeing—a critical gap our work explicitly addresses.

\begin{table}[t]
\centering
\small
\caption{Comparison between RefereeBench and representative sports datasets. RefereeBench is the first benchmark dedicated to sports refereeing across 10+ sport types. A.S. R.G. C.S.S}
\resizebox{\columnwidth}{!}{
\begin{tabular}{lccccc}
\toprule
\textbf{Dataset} & \textbf{Task} & \textbf{\# Sports} & \textbf{\# Videos} & \textbf{\# QAs} \\
\midrule
SoccerNet\cite{soccernet} & Action Spotting & 1 & 500 & 6,637 \\
SoccerNet-v2\cite{soccernet-v2} & Action Spotting, etc. & 1 & 500 & ~300K \\
SportQA\cite{xia2024sportqa} & Sport QA & 35 & - & ~70K  \\
Sports-QA\cite{li2024sports} & Sport QA & 8 & 6K & 94K & \\
LiveSports-3K-QA\cite{chen2025livecc} & Sport QA & 49 & 414 & 1,236 \\
SPORTU-Video\cite{xia2024sportu} & Sport QA & 7 & 1,701 & 12,048 \\
SPORTU-Text\cite{xia2024sportu} & Sport QA & 7 & - & 900 \\
SoccerNet-MVFoul\cite{held2023vars} & Refereeing & 1 & 8,293 & - \\
SoccerNet-XFoul\cite{held2024xvars} & Refereeing & 1 & 10K & 22K \\
\midrule
\textbf{RefereeBench} & \textbf{Refereeing} & \textbf{11} & \textbf{925} & \textbf{6,475} \\
\bottomrule
\end{tabular}
}

\label{tab:dataset_comparison}
\end{table}

\subsection{Automatic Sports Refereeing}

While foundational datasets focus on general sports comprehension, actual sports officiating requires highly specialized, high-stakes decision-making\cite{xusurvey}. In the real world, technology-assisted officiating is already a practical reality. Systems such as the Video Assistant Referee (VAR)\cite{ifab2024var} and Semi-Automated Offside Technology (SAOT)\cite{fifa2022saot} in football formalize how multi-view visual evidence and spatial tracking can support human officials. Motivated by these real-world systems, the multimedia and computer vision communities have begun formalizing refereeing as an academic task. Initial efforts formulated officiating as a fine-grained action recognition or spotting problem\cite{soccernet, soccernet-v2}. For instance, VARS\cite{held2023vars} targets automated foul detection in soccer from multiple camera views. X-VARS\cite{held2024xvars} extended this trajectory by utilizing Large Language Models (LLMs) to provide textual justifications for football refereeing decisions. Despite this progress, current automatic refereeing research remains severely constrained. Pioneering sports refereeing works are strictly confined to soccer. Furthermore, while broad multi-sport benchmarks like SportQA\cite{xia2024sportqa} and SPORTU\cite{xia2024sportu}include foul or penalty recognition subsets, these tasks only constitute a minor fraction of their evaluation. Consequently, the literature currently lacks a large-scale, dedicated benchmark specifically designed to stress-test the cross-sport, explainable officiating capabilities of MLLMs.



\section{RefereeBench}

\subsection{Data Construction}
The construction of RefereeBench mainly involves three steps: video collection, QA annotation, and quality control. The details are as follows:

\paragraph{\textbf{Video Collection}}
To collect officiating incidents at scale, we first source raw videos from YouTube, official competition platforms, and major sports media websites, encompassing \textbf{11 distinct sports: soccer, basketball, volleyball, tennis, table tennis, badminton, handball, field hockey, ice hockey, water polo, and short track speed skating}. In total, we amassed an extensive initial corpus of 5,410 raw videos. To ensure authentic and continuous match contexts, we prioritize full match replays and extended broadcast footage. Concurrently, we filter out unsuitable content—such as highlight compilations, post-match interviews, and heavily edited short clips—by leveraging video titles and metadata.

To localize candidate officiating incidents within these long, untrimmed videos, we leverage broadcast commentary as our primary semantic cue. We first extract existing subtitle streams or apply Automatic Speech Recognition (ASR) to the audio tracks using WhisperX\cite{bain2023whisperx} to obtain time-aligned transcripts. Next, we perform keyword matching on these transcripts using a comprehensive refereeing lexicon, which comprises both generic officiating terms (e.g., "foul", "penalty") and sport-specific rule expressions. To capture the full context of an incident, the matched timestamps are padded with predefined temporal windows to generate candidate video proposals. Through this automated process, we generated a massive set of over 40,000 candidate proposals.

Finally, these proposals undergo rigorous human review via a dedicated annotation interface. Annotators verify whether each proposal contains a valid officiating event and precisely refine its temporal boundaries. After discarding false positives and irrelevant scenes, this strict filtering process yielded a highly curated set of 925 verified segments. These segments are subsequently cropped into standalone video clips, forming the high-quality raw video corpus for the downstream QA annotation.

\paragraph{\textbf{QA Annotation}}
To ensure high-quality and reliable data, we employ a rigorous, \textbf{fully human-annotated, two-stage process} for QA construction. Rather than formulating questions directly from raw video clips, annotators first extract structured officiating metadata, which subsequently serves as the foundation for question generation. This intermediate structuring significantly reduces annotation drift and aligns subjective judgments across different sports.

Our annotation team consists of approximately \textbf{30 certified national-level referees} with strong English proficiency, ensuring professional domain expertise. At least two experts are assigned to each sport. To minimize fatigue-induced errors and guarantee high-quality labeling, annotators are permitted to review the clips repeatedly and proceed at a flexible pace.

In the first stage, annotators systematically record comprehensive structured metadata for each clip. This includes foul existence, foul type, foul rationale, penalty type, penalty rationale, incident timestamps across available views, involved entities, and their overall annotation confidence. In the second stage, relying on the established metadata, annotators construct 6 single-answer multiple-choice questions for each clip (excluding Q1), yielding a total of 6,475 QA pairs. Here, the structured labels act as the definitive ground truth, while annotators are tasked with phrasing the natural language questions and crafting plausible distractors (i.e., challenging negative options) tailored to the specific video context. These seven questions systematically cover five evaluation dimensions: \textbf{(1) foul existence (Q1)}, identifying whether there is a foul or not, \textbf{(2) classification (Q2 and Q4)}, targeting foul type and penalty type, respectively, \textbf{(3) reasoning (Q3 and Q5)}, assessing the rationales for the foul and penalty, \textbf{(4) entity perception (Q6)}, identifying involved players, and \textbf{(5) temporal grounding (Q7)}, pinpointing the incident timestamp or intervals. For Q1, we automatically construct from annotations using the template "Is there a foul or violation in the video or not?" for questions and "A: Yes, there is.", "B: No, there isn't" and "C: I am not Sure." for options.

We adopt a multiple-choice format rather than free-form generation to guarantee evaluation stability and reproducibility. Relying on free-form answers frequently necessitates an LLM-based evaluator, which can inherently introduce secondary biases. Notably, we are the first to introduce temporal grounding into automated sports refereeing tasks. We argue that a reliable AI referee assistant must be capable of pinpointing exactly when a foul occurs or is about to unfold. This temporal awareness is crucial because precise localization not only provides the exact visual evidence required for human review (e.g., VAR interventions) but also ensures that the model's decision is strictly anchored to the specific incident rather than spurious contextual correlations.

\paragraph{\textbf{Quality Control}}

To guarantee the quality of our dataset, we conduct a rigorous manual review process. First, for metadata annotation, we employ a cross-verification protocol. Each clip is independently reviewed by a second expert annotator. If the two sets of labels align, they are accepted; otherwise, the instance undergoes a formal adjudication process involving a senior referee until a consensus is reached. This step minimizes subjective bias and ensures a high degree of consistency across diverse sports. Second, for QA construction, every question-answer pair undergoes a peer-review stage by an annotator other than the original author. Reviewers evaluate whether the question is phrased clearly, and crucially, whether the video provides sufficient evidence to support the correct answer. Any items failing these criteria are returned for revision.

Furthermore, to mitigate language-only bias, we screen for QA pairs that might be solvable without visual context. We present the textual questions alone to Gemini-3-Flash to identify cases where correct answers can be inferred through linguistic shortcuts or prior knowledge. Questions that the model answers correctly in this blind setting are manually re-examined and revised to ensure they are strictly video-dependent. Under this text-only condition, the model achieves an accuracy of less than 30\%, which is close to the random-guess baseline, confirming that the benchmark remains strongly grounded in genuine multimodal evidence.

\subsection{Data Statistics}
We summarize the key statistical properties of RefereeBench in terms of video meta information, QA-pair composition, and qualitative examples as follows:

\paragraph{\textbf{Video and Meta Information}}

RefereeBench contains 925 final videos across 11 sports, each paired with its corresponding audio track. The upper-right part of Figure~2 shows the duration distribution of the collected videos. The average video duration is 37.25 seconds. Most videos fall into the medium-duration range (20–60 seconds), with 607 samples, while the dataset also includes 197 short videos (< 20 seconds) and 121 long videos (> 60 seconds). This distribution indicates that RefereeBench is primarily composed of temporally compact yet decision-complete videos, while still preserving a substantial number of longer cases that require broader temporal context.

The lower-right part of Figure~2 shows the number of videos and their average duration across sports. The number of videos ranges from 29 to 110 per sport, with an average of 84.1 videos across the 11 sports. The average duration ranges from 9.65 seconds to 67.52 seconds, reflecting substantial cross-sport variation in temporal structure and evidential context. In addition, the final videos are annotated with temporal segments indicating the relevant officiating events. Across the benchmark, the number of annotated temporal segments ranges from 29 to 222 per sport, with an average of 1.55 temporal segments per video. This indicates that while each final video is retained as a complete annotation unit, many videos contain more than one officiating-relevant segment. The left part of Figure~2 summarizes the 11 sports together with their foul and penalty taxonomies. In terms of officiating semantics, RefereeBench covers 64 unique foul types and 34 unique penalty types in total. This coverage spans the major and frequent foul and penalty categories across the included sports, indicating that the benchmark captures broad diversity in real-world officiating events and sanction mechanisms.

\begin{figure*}[htbp]
  \includegraphics[width=\textwidth]{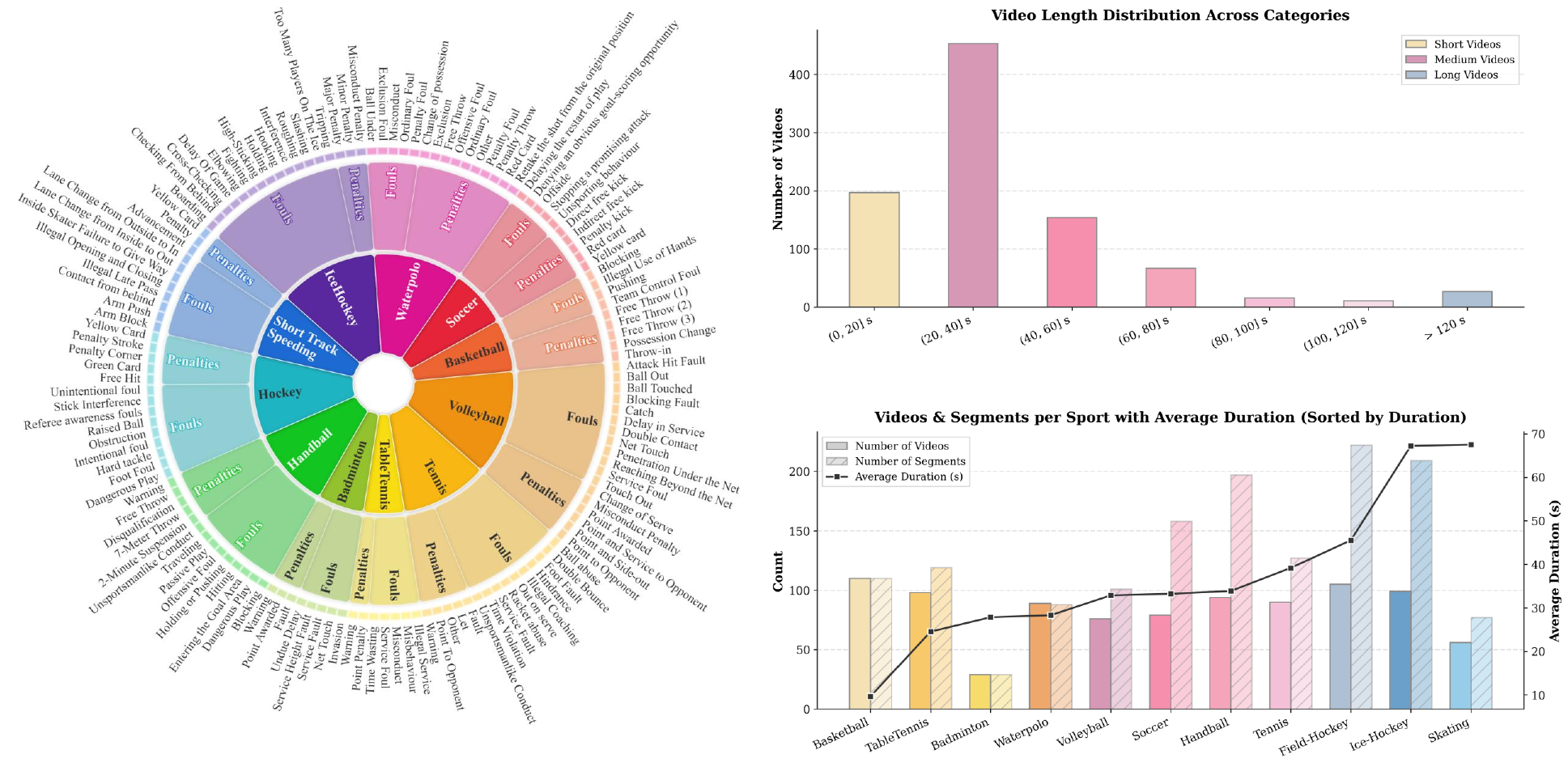}
  \caption{Statistics analysis of RefereeBench. \textit{(Left)} Sports categories and rule taxonomies. Our benchmark covers 11 distinct sports, encompassing 64 unique foul types and 34 unique penalty types. \textit{(Right)} Video duration and sample distributions across sports. RefereeBench has a full spectrum of video lengths and covers diverse officiating semantics, enabling a comprehensive evaluation of automatic sports refereeing.}
  \label{fig:statistic}
\end{figure*}

\begin{table}[t]
\centering
\caption{Linguistic statistics of QA annotations in RefereeBench. Statistics are computed after removing stopwords and punctuation.}
\label{tab:qa_linguistic_stats}
\begin{tabular}{lccc}
\toprule
\textbf{Statistic} & \textbf{Question} & \textbf{Options} & \textbf{Rationale} \\
\midrule
Word Count       & 42,633  & 119,231 & 335,450 \\
Vocabulary Size  & 509     & 3,510   & 5,888   \\
Average Length   & 13.3    & 7.6     & 89.6    \\
\bottomrule
\end{tabular}
\end{table}

\paragraph{\textbf{QA Pairs}}

RefereeBench contains 6,475 QA pairs in total, covering 11 sports. Since each final video is paired with the fixed question template Q1–Q7, the QA distribution is directly aligned with the video distribution across sports, with an average of 588.6 QA pairs across the 11 sports.

The benchmark is organized into seven question types that systematically cover five evaluation dimensions: (1) foul existence (Q1), identifying whether there is a foul or not; (2) classification (Q2 and Q4), targeting foul type and penalty type, respectively; (3) reasoning (Q3 and Q5), assessing the rationales for the foul and penalty; (4) entity perception (Q6), identifying the involved players; and (5) temporal grounding (Q7), pinpointing the incident timestamp or interval. In terms of answer format, Q1 is a 3-choice question, while Q2–Q7 are 4-choice questions. For Q2–Q7, the correct-answer options A/B/C/D follow a near-uniform distribution (25.6\% / 24.4\% / 26.3\% / 23.7\%), reducing positional bias in evaluation.

Table~\ref{tab:qa_linguistic_stats} summarizes the linguistic statistics of the QA annotations. After removing stopwords and punctuation, the annotations contain 42,633 content words in questions, 119,231 in answer options, and 335,450 in rationales, with corresponding vocabulary sizes of 509, 3,510, and 5,888. Questions remain concise, with an average length of 13.3 tokens, while answer options average 7.6 tokens. By contrast, rationales are substantially longer, with an average length of 89.6 tokens. These statistics indicate that RefereeBench combines concise and standardized question formulation with substantially richer explanatory content, making the benchmark both evaluation-friendly and semantically informative.

\paragraph{\textbf{Qualitative Analysis}}

Building on the above statistics, RefereeBench is both diverse and challenging, making it a realistic testbed for automatic sports refereeing. Figure~1 presents two representative examples from our dataset.

In the first example, taken from ice hockey, the model must distinguish the correct foul type from several visually similar candidates and identify the specific contact pattern that directly triggers the referee’s intervention. This requires not only recognizing the action itself, but also isolating the legally decisive aspect of the incident from fine-grained temporal evidence. In the second example, from soccer, the model must determine the correct penalty and identify the rule principle that best supports the decision. Here, simple incident detection is insufficient: the model must connect the observed challenge to the appropriate disciplinary standard and justify why one rule-based interpretation is correct while others are not.

These examples highlight that RefereeBench is not limited to coarse event recognition. Instead, it requires models to align visual evidence with sport-specific officiating rules and make structured decisions across different sports, thereby providing a challenging benchmark for rule-grounded multimodal reasoning.

\begin{table*}[htbp]
    \centering
    \caption{Performance of various MLLMs across 5 ability categories in the RefereeBench benchmark, with all values averaged across different sports. Best results within each model group are highlighted in bold.}
    \resizebox{\textwidth}{!}{
    \begin{tabular}{lcc ccc ccc ccc}
        \toprule
        \multirow{2}{*}{\textbf{Models}} & \multirow{2}{*}{\textbf{Params}} & \textbf{Existence (\%)} & \multicolumn{3}{c}{\textbf{Classification (\%)}} & \multicolumn{3}{c}{\textbf{Reasoning (\%)}} & \textbf{Perception (\%)} & \textbf{Grounding (\%)} & \multirow{2}{*}{\textbf{OVERALL (\%)}} \\
        \cmidrule(lr){3-3} \cmidrule(lr){4-6} \cmidrule(lr){7-9} \cmidrule(lr){10-10} \cmidrule(lr){11-11}
        & & \textbf{Q1} & \textbf{Q2} & \textbf{Q4} & \textbf{Avg.} & \textbf{Q3} & \textbf{Q5} & \textbf{Avg.} & \textbf{Q6} & \textbf{Q7} & \\
        \midrule
        \multicolumn{12}{c}{\textit{Closed-source MLLMs}} \\
        \midrule
        GPT-5\cite{openai_gpt5_2025} & - & 64.2 & 54.1 & 59.1 & 56.6 & 48.2 & 47.4 & 47.8 & 72.6 & 38.1 & 54.8 \\
        GPT-4o\cite{openai_gpt4o_2024} & - & 63.0 & 44.4 & 49.7 & 47.1 & 38.8 & 39.5 & 39.1 & 65.3 & 28.6 & 47.1 \\
        Gemini-3-Pro\cite{gemini3_google_2025} & - & 71.0 & 54.3 & 61.5 & 57.9 & 50.6 & 55.3 & 53.0 & 78.6 & 48.8 & 60.0 \\
        Gemini-3-Flash\cite{gemini3_google_2025} & - & 74.2 & 57.2 & 60.3 & \textbf{58.8} & 50.5 & 55.6 & \textbf{53.1} & \textbf{79.2} & 46.8 & 60.6 \\
        Claude-4.5-Sonnet\cite{claude45_anthropic_2025} & - & 65.4 & 38.8 & 51.9 & 45.4 & 42.4 & 42.5 & 42.5 & 59.1 & 33.3 & 47.6 \\
        Claude-4.5-Haiku\cite{claude45_anthropic_2025} & - & 81.0 & 32.1 & 38.7 & 35.4 & 37.1 & 40.4 & 38.7 & 55.4 & 34.5 & 45.6 \\
        Doubao-Seed-1.8\cite{bytedance2025seed18} & - & \textbf{92.5} & 49.6 & 57.1 & 53.4 & 50.9 & 50.5 & 50.7 & 76.8 & \textbf{50.1} & \textbf{61.1} \\
        \midrule
        \multicolumn{12}{c}{\textit{Open-source \& Video MLLMs}} \\
        \midrule
        Qwen3-VL\cite{bai2025qwen3} & 8B & \textbf{50.8} & 41.4 & 56.7 & \textbf{49.0} & 39.9 & 34.4 & \textbf{37.1} & \textbf{67.4} & \textbf{39.0} & \textbf{47.1} \\
        InternVL-3.5\cite{wang2025internvl35} & 8B & 43.9 & 35.8 & 47.4 & 41.6 & 39.1 & 35.1 & \textbf{37.1} & 61.8 & 29.6 & 41.8 \\
        VideoLLaMA3\cite{zhang2025videollama} & 7B & 33.2 & 31.6 & 38.0 & 34.8 & 32.7 & 31.8 & 32.3 & 54.8 & 30.1 & 36.0 \\
        LLaVA-Video\cite{zhang2024llava} & 7B & 37.2 & 23.7 & 26.5 & 25.1 & 23.7 & 26.1 & 24.9 & 26.7 & 27.8 & 27.4 \\
        \bottomrule
    \end{tabular}
    }

    \label{tab:hierarchical_abilities}
\end{table*}

\begin{figure*}
  \includegraphics[width=0.95\textwidth]{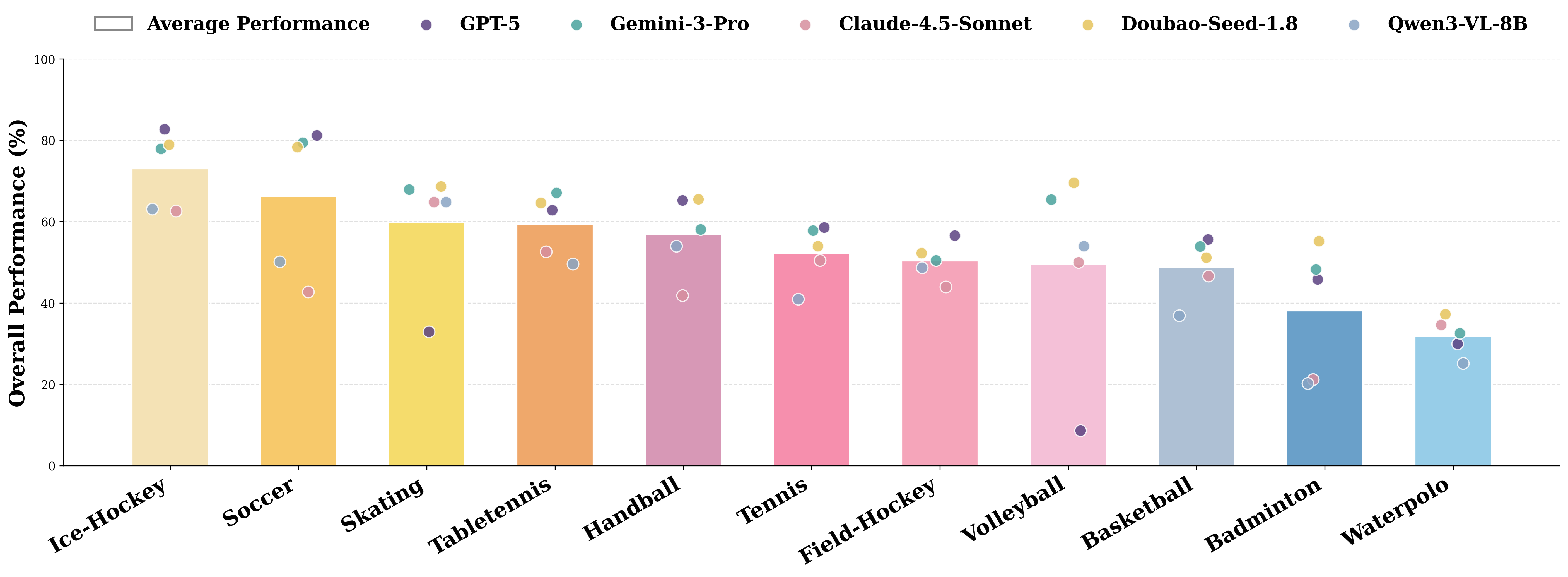}
  \caption{Overall performance on RefereeBench of representative MLLMs across different sports. Results are presented in descending order of average accuracy across categories. The results highlight that automated refereeing difficulty and model generalization vary substantially depending on the specific sport.}
  \label{fig:performance_across_sports}
\end{figure*}

\section{Experiments}

In this section, we evaluate various MLLMs on our RefereeBench. We begin by introducing the experiment settings and showing the quantitative results for both closed-source and open-source models. Then, we provide a deep analysis of the model results by analyzing what the bottleneck is for models to act like real-world human referees.

\subsection{Settings}

We evaluate RefereeBench on a diverse set of proprietary and open-source MLLMs. For closed-source commercial models, we include GPT-5\cite{openai_gpt5_2025}, GPT-4o\cite{openai_gpt4o_2024}, Gemini 3 Pro\cite{gemini3_google_2025}, Gemini 3 Flash\cite{gemini3_google_2025}, Claude Sonnet 4.5\cite{claude45_anthropic_2025}, Claude Haiku 4.5\cite{claude45_anthropic_2025}, and Doubao Seed 1.8\cite{bytedance2025seed18}. The open-source video models include Qwen3-VL-8B\cite{bai2025qwen3}, InternVL3.5-8B\cite{wang2025internvl35}, VideoLLaMA3-7B\cite{zhang2025videollama}, and LLaVA-Video-7B\cite{zhang2024llava}. For all models, we follow their official configurations and try to use more frames for evaluation. All models are evaluated under a unified multiple-choice answering protocol, and each question is presented together with its corresponding full video. We follow all default FPS settings for each model, and the video resolution is fixed at 720p. Details of each model's settings can be found in the Appendix. The accuracy is computed by directly comparing the model’s output with the ground truth answer without the need for any external models. 

\begin{table*}[htbp]
    \centering
    \caption{Effect of audio on RefereeBench. We evaluate Gemini-3-Flash and Qwen3-Omni under two settings: frame-only input and direct video-file input with audio. Adding audio consistently improves performance.}
    \resizebox{\textwidth}{!}{
    \begin{tabular}{ll c ccc ccc c c c}
        \toprule
        \multirow{2}{*}{\textbf{Models}} & \multirow{2}{*}{\textbf{Modality}} & \textbf{Existence (\%)} & \multicolumn{3}{c}{\textbf{Classification (\%)}} & \multicolumn{3}{c}{\textbf{Reasoning (\%)}} & \textbf{Perception (\%)} & \textbf{Grounding (\%)} & \multirow{2}{*}{\textbf{OVERALL (\%)}} \\
        \cmidrule(lr){3-3} \cmidrule(lr){4-6} \cmidrule(lr){7-9} \cmidrule(lr){10-10} \cmidrule(lr){11-11}
        & & \textbf{Q1} & \textbf{Q2} & \textbf{Q4} & \textbf{Avg.} & \textbf{Q3} & \textbf{Q5} & \textbf{Avg.} & \textbf{Q6} & \textbf{Q7} & \\
        \midrule
        \multirow{2}{*}{Gemini-3-Flash} & Frames & 74.2 & 57.2 & 60.3 & 58.8 & 50.5 & 55.6 & 53.1 & 79.2 & 46.8 & 60.6 \\
        & + Audio & \textbf{96.8} & \textbf{76.9} & \textbf{79.0} & \textbf{78.0} & \textbf{63.6} & \textbf{73.4} & \textbf{68.5} & \textbf{81.8} & \textbf{64.3} & \textbf{76.5} \gain{16.0} \\
        \midrule
        \multirow{2}{*}{Qwen3-Omni\cite{xu2025qwen3}} & Frames & 56.7 & 45.6 & 57.3 & 51.5 & 42.8 & 41.7 & 42.2 & 66.3 & 32.5 & 49.0 \\
        & + Audio & \textbf{87.0} & \textbf{62.8} & \textbf{73.7} & \textbf{68.3} & \textbf{54.2} & \textbf{55.2} & \textbf{54.7} & \textbf{74.7} & \textbf{51.8} & \textbf{65.7} \gain{16.7} \\
        \bottomrule
    \end{tabular}
    }
    \label{tab:hierarchical_abilities_new}
\end{table*}

\subsection{Quantitative Results}

\paragraph{\textbf{Performance of Closed-Source Models.}}
Table~\ref{tab:hierarchical_abilities} reports the main results on RefereeBench. Among the closed-source models, Doubao-Seed-1.8 achieves the best overall accuracy of 61.09\%, making it the strongest model on the benchmark. The two Gemini models perform very competitively, with Gemini-3-Flash and Gemini-3-Pro reaching 60.55\% and 60.02\%, respectively, only marginally below Doubao-Seed-1.8. By contrast, the GPT series lags behind this top group: GPT-5 remains relatively competitive at 54.82\%, whereas GPT-4o drops to 47.06\%. The Claude models perform less favorably on this task, with Claude-4.5-Sonnet and Claude-4.5-Haiku obtaining only 47.64\% and 45.59\%, respectively. These results show that, despite their strong general multimodal capabilities, even leading commercial systems remain far from reliable automatic referees.

A closer look at the ability breakdown reveals a clear pattern. Closed-source models are strongest on Existence and Perception, suggesting that they are relatively effective at recognizing whether an incident occurs and identifying the involved entities. For example, Doubao-Seed-1.8 reaches 92.54\% on Q1 and 76.76\% on Q6, while the two Gemini models achieve around 79\% on Q6. However, Reasoning and Grounding remain substantially harder. Even the strongest closed-source models only reach the low-50\% range on the aggregated Reasoning category, and their Q7 scores remain around or below 50\%. This indicates that incident detection is considerably more tractable than applying sport-specific rules to concrete game situations and pinpointing the decisive temporal evidence.

\paragraph{\textbf{Performance of Open-Source Models.}}
The open-source models exhibit a clear performance gap relative to the strongest closed-source systems. Qwen3-VL is the strongest open-source baseline with an overall accuracy of 47.07\%, followed by InternVL3.5 at 41.82\% and VideoLLaMA3 at 36.04\%. LLaVA-Video achieves only 27.40\%, which is close to random performance. This highlights the substantial difficulty of RefereeBench for current open-source video MLLMs and underscores the clear gap between open and closed-source models on referee-oriented reasoning.

Although the strongest open-source model demonstrates non-trivial competence, it still remains well below the best closed-source models. In particular, Qwen3-VL trails Doubao-Seed-1.8 by more than 14 percentage points overall. Results also suggest that current open models retain some ability to capture coarse event structure, but remain much less reliable when judgments require deeper rule application. For instance, Qwen3-VL performs relatively better on Perception (67.35\%) and reaches 50.81\% on Existence, yet drops to only 37.14\% on aggregated Reasoning and 38.96\% on Grounding, showing that open-source MLLMs fail to transform perceptual evidence into stable, rule-consistent decisions.

\paragraph{\textbf{Performance Across Sports.}}
Results in Figure~\ref{fig:performance_across_sports} show that model performance varies substantially across sports. This variation is observed across both closed-source and open-source models, suggesting that sport-level difficulty is a persistent challenge rather than a model-specific artifact. In particular, some sports are consistently easier for most models (e.g. Ice-Hockey and Soccer), whereas others remain challenging across nearly all systems (e.g. Badminton and Waterpolo). Another notable observation is that strong overall performance does not necessarily imply stable cross-sport generalization. A model that performs well on average may still exhibit clear weaknesses on particular sports, and the relative ranking of models is not perfectly consistent across sports. This variation may reflect both training-data bias across sports and differences in the visual characteristics and rule structures of the sports themselves. These results indicate that RefereeBench captures meaningful sport-level variation beyond a single global notion of difficulty.


\subsection{Analysis}

We conduct further analysis to explore the factors influencing the sports refereeing performance.

\paragraph{\textbf{Could additional modalities benefit the performance?}}
Sports refereeing is inherently multimodal, as officiating decisions often depend not only on visual evidence but also on acoustic cues. To assess the contribution of audio, we evaluate Gemini-3-Flash and Qwen3-Omni under two settings: \textit{Frames}, which uses only sampled video frames, and \textit{+ Audio}, which directly uploads the original video file for inference. Table~\ref{tab:hierarchical_abilities_new} shows that adding audio substantially improves performance for both models. Gemini-3-Flash increases from 60.55\% to 76.54\%, and Qwen3-Omni increases from 48.97\% to 65.65\%, yielding gains of 15.99 and 16.68 points, respectively. The gains are consistent across all five ability categories, with the largest improvements appearing in Classification, Reasoning, and Grounding. These results suggest that audio provides meaningful evidence and that sports refereeing is inherently a multimodal reasoning task.

\begin{table}[htbp]
\centering
\small
\caption{Performance of Gemini-3-Flash and Qwen3-VL on text-only professional referee exam questions, averaged across sports. Rule, Mgmt., and Exec. denote Rule Knowledge, Game Management, and Officiating Execution, respectively.}
\label{tab:referee_exam_merged}
\resizebox{\columnwidth}{!}{
\begin{tabular}{lcccc}
\toprule
\textbf{Model} & \textbf{Rule (\%)} & \textbf{Mgmt. (\%)} & \textbf{Exec. (\%)} & \textbf{OVERALL (\%)} \\
\midrule
Gemini-3-Flash & 70.5 & 58.4 & 61.1 & \textbf{69.3} \\
Qwen3-VL-8B   & 54.5  & 44.6   & 49.0  & \textbf{53.7}   \\
\bottomrule
\end{tabular}
}
\end{table}

\paragraph{\textbf{How well do models perform on sports refereeing knowledge?}}
To assess sports-domain knowledge for refereeing, we further evaluate MLLMs in a text-only setting using professional referee exam questions collected from the web with gold-standard answers. In total, we collect \textbf{2,684} questions across \textbf{11 sports}. Following international refereeing competency standards, we organize the text-based QAs into three dimensions: \textbf{(1) Rule Knowledge}, covering the understanding of rules, specifications, and officiating protocols; \textbf{(2) Game Management}, covering the handling of player behavior, communication, and match situations; and \textbf{(3) Officiating Execution}, covering practical officiating skills such as mechanics, positioning, and signals. More details are provided in the Appendix. As shown in Table~\ref{tab:referee_exam_merged}, Gemini-3-Flash achieves \textbf{69.3\%} overall accuracy, clearly outperforming Qwen3-VL-8B at \textbf{53.7\%}. This trend is also consistent with the main benchmark results, where Gemini-3-Flash maintains a clear advantage over Qwen3-VL-8B. For both models, Rule Knowledge is the strongest category, whereas Game Management is the weakest. These results indicate that current MLLMs possess non-trivial but incomplete refereeing knowledge, and that the challenge of automatic sports refereeing arises not only from multimodal grounding, but also from limited sports-domain knowledge itself~\cite{xia2024sportqa, xia2024sportu}.

\begin{table}[t]
\centering
\small
\caption{Vulnerability to suggestive bias on negative samples. We report the misidentification rate (\%) under neutral and suggestive settings. Lower is better.}
\label{tab:suggestive_bias}
\resizebox{\columnwidth}{!}{
\begin{tabular}{l c|cc|cc}
\toprule
\multirow{2}{*}{\textbf{Model}} & \textbf{Q1 (\%)} & \multicolumn{2}{c|}{\textbf{Q2 (Foul) (\%)}} & \multicolumn{2}{c}{\textbf{Q4 (Penalty) (\%)}} \\
\cmidrule(lr){2-2} \cmidrule(lr){3-4} \cmidrule(lr){5-6}
& \textbf{neutral} & \textbf{neutral} & \textbf{Sugg.} & \textbf{neutral} & \textbf{Sugg.} \\
\midrule
Gemini-3-Flash & 37.7 & 63.6 & 76.8 & 57.7 & 76.8 \\
Qwen3-VL-8B    & 12.7 & 41.6 & 60.4 & 56.5 & 74.9 \\
\bottomrule
\end{tabular}
}
\end{table}

\paragraph{\textbf{How vulnerable are MLLMs on negative samples?}}
We conduct this experiment to examine whether current MLLMs tend to over-call fouls on negative samples, i.e., normal sports clips that do not contain an actual violation. To this end, we collect negative samples using the same method as the positive set, with 20 videos for each sport. We focus on Q1, Q2, and Q4, corresponding to foul existence, foul classification, and penalty classification, respectively. During question design, we explicitly ensure that each question contains a valid negative option (such as \textit{no}, \textit{no foul}, or \textit{no penalty}).

Table~\ref{tab:suggestive_bias} reports the misidentification rates under neutral and suggestive settings. Gemini-3-Flash already shows severe errors under the neutral setting, reaching 37.7\% on Q1, 63.6\% on Q2, and 57.7\% on Q4. Although Qwen3-VL-8B performs better on Q1 (12.7\%), it still exhibits high misidentification rates on Q2 and Q4, rising to 60.4\% and 74.9\% under suggestive settings. These results show that current MLLMs struggle to distinguish legal sports actions from true fouls and tend to over-officiate, especially under suggestive wording. Analysis of generated rationales further suggests different failure mechanisms across models. For example, Gemini often over-interprets legal physical contact and hallucinates specific but non-existent events, such as inventing a referee penalty signal for a normal body check.  Overall, the findings suggest that current MLLMs are still far from being able to serve as human-like sports referees, as they remain unable to make reliable and unbiased judgments.


\section{Limitations and Future Work}

In this section, we discuss current limitations of automatic sports refereeing and outline promising future directions from both data and method perspectives.

\subsection{Data}

A key limitation lies in the scale and diversity of refereeing data. Existing datasets are still biased toward a small number of popular sports\cite{xusurvey}, especially soccer, while different sports exhibit substantially different rule systems and interaction patterns. Although RefereeBench extends coverage to 11 sports, future work should further expand to a broader range of disciplines to better evaluate cross-sport generalization.

Another important challenge is the gap between benchmark data and real-world deployment. Current datasets are mostly constructed from broadcast videos, which often contain auxiliary cues such as commentator descriptions or referee gestures. These signals may introduce shortcuts that do not reflect true decision-making ability. Future work could reduce such bias by removing referee-related cues or collecting cleaner real-world data\cite{ifab2024var}.

\subsection{Methods}

We conduct preliminary explorations with inference-time enhancements and basic retrieval-augmented generation (RAG), as detailed in the Appendix, but none yields satisfactory performance, underscoring the difficulty of equipping MLLMs with reliable automatic refereeing abilities.

One important future direction is to develop stronger training-based methods. Current models still struggle to align visual evidence with sport-specific rules and decision criteria\cite{abootorabi2025ask, zhao2023retrieving}. A natural direction is to provide stronger sports-specific reasoning supervision, for example through intermediate reasoning steps or visual chain-of-thought training\cite{shao2024visual}. Prior work also shows that explanation-oriented supervision can improve refereeing reasoning and interpretability\cite{held2024xvars}. In addition, models tend to over-call fouls, suggesting a need for better decision calibration. Learning from real-world preference data and applying reinforcement learning may help encourage more conservative and reliable judgments\cite{rafailov2023direct, schulman2017proximal, shao2024deepseekmath}.

Beyond training, future work may explore stronger RAG frameworks that dynamically retrieve relevant clauses from sport rulebooks and explicitly ground decisions in them\cite{li2024benchmarking}. Adopting agentic pipelines for refereeing can also be beneficial. Instead of a single-step prediction, the task can be decomposed into stages such as incident detection, candidate generation, and rule verification, leading to more stable and interpretable outputs. Similar advantages have been observed in prior work on general sport understanding\cite{rao2025multi}.

\section{Conclusion}

In this paper, we introduced RefereeBench, the first comprehensive benchmark for evaluating MLLMs on multi-sport automatic refereeing. Covering 11 sports and five hierarchical dimensions, RefereeBench provides a rigorous testbed for complex multimodal decision-making. Our extensive evaluations reveal a clear gap in current MLLM capabilities: although models show basic perceptual competence, they still struggle with precise temporal localization and sport-specific rule application. These results show that current MLLMs are not yet ready to function as reliable multi-sport referees, as they still struggle to apply precise sport-specific knowledge to specific game scenarios and remain highly vulnerable to suggestive bias. We hope RefereeBench can encourage the multimedia community to move beyond generic video understanding toward trustworthy, expert-level AI assistants with robust, rule-grounded reasoning in high-stakes real-world environments.

\bibliographystyle{ACM-Reference-Format}
\bibliography{assets/reference}

@String{Computer = "{IEEE} Computer" }

@String{Chelsea = "Chelsea" }

@inproceedings{soccernet,
  title={Soccernet: A scalable dataset for action spotting in soccer videos},
  author={Giancola, Silvio and Amine, Mohieddine and Dghaily, Tarek and Ghanem, Bernard},
  booktitle={Proceedings of the IEEE conference on computer vision and pattern recognition workshops},
  pages={1711--1721},
  year={2018}
}

@inproceedings{soccernet-v2,
  title={Soccernet-v2: A dataset and benchmarks for holistic understanding of broadcast soccer videos},
  author={Deliege, Adrien and Cioppa, Anthony and Giancola, Silvio and Seikavandi, Meisam J and Dueholm, Jacob V and Nasrollahi, Kamal and Ghanem, Bernard and Moeslund, Thomas B and Van Droogenbroeck, Marc},
  booktitle={Proceedings of the IEEE/CVF conference on computer vision and pattern recognition},
  pages={4508--4519},
  year={2021}
}

@inproceedings{soccernet-caption,
  title={SoccerNet-caption: Dense video captioning for soccer broadcasts commentaries},
  author={Mkhallati, Hassan and Cioppa, Anthony and Giancola, Silvio and Ghanem, Bernard and Van Droogenbroeck, Marc},
  booktitle={Proceedings of the IEEE/CVF Conference on Computer Vision and Pattern Recognition},
  pages={5074--5085},
  year={2023}
}

@inproceedings{qi2023goal,
  title={GOAL: A challenging knowledge-grounded video captioning benchmark for real-time soccer commentary generation},
  author={Qi, Ji and Yu, Jifan and Tu, Teng and Gao, Kunyu and Xu, Yifan and Guan, Xinyu and Wang, Xiaozhi and Xu, Bin and Hou, Lei and Li, Juanzi and others},
  booktitle={Proceedings of the 32nd ACM international conference on information and knowledge management},
  pages={5391--5395},
  year={2023}
}

@inproceedings{gao2025fsbench,
  title={Fsbench: A figure skating benchmark for advancing artistic sports understanding},
  author={Gao, Rong and Liu, Xin and Hu, Zhuozhao and Xing, Bohao and Xia, Baiqiang and Yu, Zitong and K{\"a}lvi{\"a}inen, Heikki},
  booktitle={Proceedings of the Computer Vision and Pattern Recognition Conference},
  pages={13595--13605},
  year={2025}
}

@inproceedings{shao2020finegym,
  title={Finegym: A hierarchical video dataset for fine-grained action understanding},
  author={Shao, Dian and Zhao, Yue and Dai, Bo and Lin, Dahua},
  booktitle={Proceedings of the IEEE/CVF conference on computer vision and pattern recognition},
  pages={2616--2625},
  year={2020}
}

@inproceedings{xu2022finediving,
  title={Finediving: A fine-grained dataset for procedure-aware action quality assessment},
  author={Xu, Jinglin and Rao, Yongming and Yu, Xumin and Chen, Guangyi and Zhou, Jie and Lu, Jiwen},
  booktitle={Proceedings of the IEEE/CVF conference on computer vision and pattern recognition},
  pages={2949--2958},
  year={2022}
}

@article{ge2024scbench,
  title={Scbench: A sports commentary benchmark for video LLMs},
  author={Ge, Kuangzhi and Chen, Lingjun and Zhang, Kevin and Luo, Yulin and Shi, Tianyu and Fan, Liaoyuan and Li, Xiang and Wang, Guanqun and Zhang, Shanghang},
  journal={arXiv preprint arXiv:2412.17637},
  year={2024}
}

@inproceedings{you2025timesoccer,
  title={Timesoccer: An end-to-end multimodal large language model for soccer commentary generation},
  author={You, Ling and Huang, Wenxuan and Xie, Xinni and Wei, Xiangyi and Li, Bangyan and Lin, Shaohui and Li, Yang and Wang, Changbo},
  booktitle={Proceedings of the 33rd ACM International Conference on Multimedia},
  pages={3418--3427},
  year={2025}
}

@article{tian2025sportsgpt,
  title={SportsGPT: An LLM-driven Framework for Interpretable Sports Motion Assessment and Training Guidance},
  author={Tian, Wenbo and Lin, Ruting and Zheng, Hongxian and Yang, Yaodong and Wu, Geng and Zhang, Zihao and Zhang, Zhang},
  journal={arXiv preprint arXiv:2512.14121},
  year={2025}
}

@inproceedings{rao2025multi,
  title={Multi-agent system for comprehensive soccer understanding},
  author={Rao, Jiayuan and Li, Zifeng and Wu, Haoning and Zhang, Ya and Wang, Yanfeng and Xie, Weidi},
  booktitle={Proceedings of the 33rd ACM International Conference on Multimedia},
  pages={3654--3663},
  year={2025}
}

@inproceedings{diaz2025soccerhigh,
  title={SoccerHigh: a benchmark dataset for automatic soccer video summarization},
  author={D{\'\i}az-Juan, Artur and Ballester, Coloma and Haro, Gloria},
  booktitle={Proceedings of the 8th International ACM Workshop on Multimedia Content Analysis in Sports},
  pages={121--130},
  year={2025}
}

@article{wang2024tacticai,
  title={TacticAI: an AI assistant for football tactics},
  author={Wang, Zhe and Veli{\v{c}}kovi{\'c}, Petar and Hennes, Daniel and Toma{\v{s}}ev, Nenad and Prince, Laurel and Kaisers, Michael and Bachrach, Yoram and Elie, Romuald and Wenliang, Li Kevin and Piccinini, Federico and others},
  journal={Nature communications},
  volume={15},
  number={1},
  pages={1906},
  year={2024},
  publisher={Nature Publishing Group UK London}
}

@inproceedings{honda2022pass,
  title={Pass receiver prediction in soccer using video and players' trajectories},
  author={Honda, Yutaro and Kawakami, Rei and Yoshihashi, Ryota and Kato, Kenta and Naemura, Takeshi},
  booktitle={Proceedings of the IEEE/CVF conference on computer vision and pattern recognition},
  pages={3503--3512},
  year={2022}
}

@article{xusurvey,
  title={A Survey of Large Models in Sports},
  author={Xu, Yichen and Ma, Jianzhe and Wang, Chuhan and Cao, Zhonghao and Chen, Liangyu and Wang, Wenxuan and Jin, Qin}
}

@article{xia2024language,
  title={Language and multimodal models in sports: A survey of datasets and applications},
  author={Xia, Haotian and Yang, Zhengbang and Zhao, Yun and Wang, Yuqing and Li, Jingxi and Tracy, Rhys and Zhu, Zhuangdi and Wang, Yuan-fang and Chen, Hanjie and Shen, Weining},
  journal={arXiv preprint arXiv:2406.12252},
  year={2024}
}

@article{zhao2025survey,
  title={A survey of deep learning in sports applications: Perception, comprehension, and decision},
  author={Zhao, Zhonghan and Chai, Wenhao and Hao, Shengyu and Hu, Wenhao and Wang, Guanhong and Cao, Shidong and Song, Mingli and Hwang, Jenq-Neng and Wang, Gaoang},
  journal={IEEE Transactions on Visualization and Computer Graphics},
  year={2025},
  publisher={IEEE}
}

@inproceedings{song2025video,
  title={Video-mmlu: A massive multi-discipline lecture understanding benchmark},
  author={Song, Enxin and Chai, Wenhao and Xu, Weili and Xie, Jianwen and Liu, Yuxuan and Wang, Gaoang},
  booktitle={Proceedings of the IEEE/CVF International Conference on Computer Vision},
  pages={6099--6113},
  year={2025}
}

@inproceedings{li2024mvbench,
  title={Mvbench: A comprehensive multi-modal video understanding benchmark},
  author={Li, Kunchang and Wang, Yali and He, Yinan and Li, Yizhuo and Wang, Yi and Liu, Yi and Wang, Zun and Xu, Jilan and Chen, Guo and Luo, Ping and others},
  booktitle={Proceedings of the IEEE/CVF Conference on Computer Vision and Pattern Recognition},
  pages={22195--22206},
  year={2024}
}

@inproceedings{fu2025video,
  title={Video-mme: The first-ever comprehensive evaluation benchmark of multi-modal llms in video analysis},
  author={Fu, Chaoyou and Dai, Yuhan and Luo, Yongdong and Li, Lei and Ren, Shuhuai and Zhang, Renrui and Wang, Zihan and Zhou, Chenyu and Shen, Yunhang and Zhang, Mengdan and others},
  booktitle={Proceedings of the IEEE/CVF conference on computer vision and pattern recognition},
  pages={24108--24118},
  year={2025}
}

@article{liu2025videoreasonbench,
  title={VideoReasonBench: Can MLLMs Perform Vision-Centric Complex Video Reasoning?},
  author={Liu, Yuanxin and Ouyang, Kun and Wu, Haoning and Liu, Yi and Sui, Lin and Li, Xinhao and Zhong, Yan and Charles, Y and Zhou, Xinyu and Sun, Xu},
  journal={arXiv preprint arXiv:2505.23359},
  year={2025}
}

@article{wu2024longvideobench,
  title={Longvideobench: A benchmark for long-context interleaved video-language understanding},
  author={Wu, Haoning and Li, Dongxu and Chen, Bei and Li, Junnan},
  journal={Advances in Neural Information Processing Systems},
  volume={37},
  pages={28828--28857},
  year={2024}
}

@article{xun2025rtv,
  title={Rtv-bench: Benchmarking mllm continuous perception, understanding and reasoning through real-time video},
  author={Xun, Shuhang and Tao, Sicheng and Li, Jungang and Shi, Yibo and Lin, Zhixin and Zhu, Zhanhui and Yan, Yibo and Li, Hanqian and Zhang, Linghao and Wang, Shikang and others},
  journal={arXiv preprint arXiv:2505.02064},
  year={2025}
}

@article{feng2025breaking,
  title={Breaking Down Video LLM Benchmarks: Knowledge, Spatial Perception, or True Temporal Understanding?},
  author={Feng, Bo and Lai, Zhengfeng and Li, Shiyu and Wang, Zizhen and Wang, Simon and Huang, Ping and Cao, Meng},
  journal={arXiv preprint arXiv:2505.14321},
  year={2025}
}

@inproceedings{held2023vars,
  title={VARS: Video assistant referee system for automated soccer decision making from multiple views},
  author={Held, Jan and Cioppa, Anthony and Giancola, Silvio and Ghanem, Bernard and Van Droogenbroeck, Marc},
  booktitle={Proceedings of the IEEE/CVF Conference on Computer Vision and Pattern Recognition Workshops},
  pages={5086--5097},
  year={2023}
}

@inproceedings{held2024xvars,
  title={X-vars: Introducing explainability in football refereeing with multi-modal large language models},
  author={Held, Jan and Itani, Hani and Cioppa, Anthony and Giancola, Silvio and Ghanem, Bernard and Van Droogenbroeck, Marc},
  booktitle={Proceedings of the IEEE/CVF conference on computer vision and pattern recognition},
  pages={3267--3279},
  year={2024}
}

@article{li2024sports,
  title={Sports-qa: A large-scale video question answering benchmark for complex and professional sports},
  author={Li, Haopeng and Deng, Andong and Ke, Qiuhong and Liu, Jun and Rahmani, Hossein and Guo, Yulan and Schiele, Bernt and Chen, Chen},
  journal={arXiv preprint arXiv:2401.01505},
  year={2024}
}

@inproceedings{chen2025livecc,
  title={Livecc: Learning video llm with streaming speech transcription at scale},
  author={Chen, Joya and Zeng, Ziyun and Lin, Yiqi and Li, Wei and Ma, Zejun and Shou, Mike Zheng},
  booktitle={Proceedings of the Computer Vision and Pattern Recognition Conference},
  pages={29083--29095},
  year={2025}
}

@inproceedings{xia2024sportqa,
  title={SportQA: A Benchmark for Sports Understanding in Large Language Models},
  author={Xia, Haotian and Yang, Zhengbang and Wang, Yuqing and Tracy, Rhys and Zhao, Yun and Huang, Dongdong and Chen, Zezhi and Zhu, Yan and Wang, Yuan-fang and Shen, Weining},
  booktitle={Proceedings of the 2024 Conference of the North American Chapter of the Association for Computational Linguistics: Human Language Technologies (NAACL)},
  year={2024}
}

@article{xia2024sportu,
  title={Sportu: A comprehensive sports understanding benchmark for multimodal large language models},
  author={Xia, Haotian and Yang, Zhengbang and Zou, Junbo and Tracy, Rhys and Wang, Yuqing and Lu, Chi and Lai, Christopher and He, Yanjun and Shao, Xun and Xie, Zhuoqing and others},
  journal={arXiv preprint arXiv:2410.08474},
  year={2024}
}

@inproceedings{cui2023sportsmot,
  title={Sportsmot: A large multi-object tracking dataset in multiple sports scenes},
  author={Cui, Yutao and Zeng, Chenkai and Zhao, Xiaoyu and Yang, Yichun and Wu, Gangshan and Wang, Limin},
  booktitle={Proceedings of the IEEE/CVF international conference on computer vision},
  pages={9921--9931},
  year={2023}
}

@inproceedings{huang2019tracknet,
  title={Tracknet: A deep learning network for tracking high-speed and tiny objects in sports applications},
  author={Huang, Yu-Chuan and Liao, I-No and Chen, Ching-Hsuan and {\.I}k, Ts{\`\i}-U{\'\i} and Peng, Wen-Chih},
  booktitle={2019 16th IEEE international conference on advanced video and signal based surveillance (AVSS)},
  pages={1--8},
  year={2019},
  organization={IEEE}
}

@inproceedings{sun2020tracknetv2,
  title={Tracknetv2: Efficient shuttlecock tracking network},
  author={Sun, Nien-En and Lin, Yu-Ching and Chuang, Shao-Ping and Hsu, Tzu-Han and Yu, Dung-Ru and Chung, Ho-Yi and {\.I}k, Ts{\`\i}-U{\'\i}},
  booktitle={2020 International Conference on Pervasive Artificial Intelligence (ICPAI)},
  pages={86--91},
  year={2020},
  organization={IEEE}
}

@inproceedings{wang2021sportssum2,
  title={Sportssum2. 0: Generating high-quality sports news from live text commentary},
  author={Wang, Jiaan and Li, Zhixu and Yang, Qiang and Qu, Jianfeng and Chen, Zhigang and Liu, Qingsheng and Hu, Guoping},
  booktitle={Proceedings of the 30th ACM International Conference on Information \& Knowledge Management},
  pages={3463--3467},
  year={2021}
}

@inproceedings{ingwersen2023sportspose,
  title={Sportspose-a dynamic 3d sports pose dataset},
  author={Ingwersen, Christian Keilstrup and Mikkelstrup, Christian M{\o}ller and Jensen, Janus N{\o}rtoft and Hannemose, Morten Rieger and Dahl, Anders Bjorholm},
  booktitle={Proceedings of the IEEE/CVF Conference on Computer Vision and Pattern Recognition},
  pages={5219--5228},
  year={2023}
}

@inproceedings{falaleev2024enhancing,
  title={Enhancing soccer camera calibration through keypoint exploitation},
  author={Falaleev, Nikolay S and Chen, Ruilong},
  booktitle={Proceedings of the 7th ACM International Workshop on Multimedia Content Analysis in Sports},
  pages={65--73},
  year={2024}
}

@inproceedings{he2025finebadminton,
  title={Finebadminton: A multi-level dataset for fine-grained badminton video understanding},
  author={He, Xusheng and Liu, Wei and Ma, Shanshan and Liu, Qian and Ma, Chenghao and Wu, Jianlong},
  booktitle={Proceedings of the 33rd ACM International Conference on Multimedia},
  pages={12776--12783},
  year={2025}
}

@article{liu2025f,
  title={F$^3$ Set: Towards Analyzing Fast, Frequent, and Fine-grained Events from Videos},
  author={Liu, Zhaoyu and Jiang, Kan and Ma, Murong and Hou, Zhe and Lin, Yun and Dong, Jin Song},
  journal={arXiv preprint arXiv:2504.08222},
  year={2025}
}

@misc{ifab2024var,
  title = {Video Assistant Referee ({VAR}) Protocol},
  author = {{The International Football Association Board (IFAB)}},
  year = {2024},
  howpublished = {\url{https://www.theifab.com/laws/latest/video-assistant-referee-var-protocol/}}
}

@misc{fifa2022saot,
  title = {Semi-automated offside technology to be used at {FIFA World Cup 2022}},
  author = {{FIFA}},
  year = {2022},
  howpublished = {\url{https://www.fifa.com/technical/media-releases/semi-automated-offside-technology-to-be-used-at-fifa-world-cup-2022-tm}}
}

@misc{gemini3_google_2025,
  title        = {A New Era of Intelligence with Gemini 3},
  author       = {{Google}},
  year         = {2025},
  month        = nov,
  howpublished = {\url{https://blog.google/products-and-platforms/products/gemini/gemini-3/}},
  note         = {Accessed: 2026-03-29}
}

@misc{claude45_anthropic_2025,
  title        = {Introducing Claude Sonnet 4.5},
  author       = {{Anthropic}},
  year         = {2025},
  month        = sep,
  howpublished = {\url{https://www.anthropic.com/news/claude-sonnet-4-5}},
  note         = {Accessed: 2026-03-29}
}

@misc{openai_gpt5_2025,
  title        = {Introducing GPT-5},
  author       = {{OpenAI}},
  year         = {2025},
  month        = aug,
  howpublished = {\url{https://openai.com/index/introducing-gpt-5/}},
  note         = {Accessed: 2026-03-29}
}

@misc{openai_gpt4o_2024,
  title        = {Hello GPT-4o},
  author       = {{OpenAI}},
  year         = {2024},
  month        = may,
  howpublished = {\url{https://openai.com/index/hello-gpt-4o/}},
  note         = {Accessed: 2026-03-29}
}

@misc{bytedance2025seed18,
  author       = {{ByteDance Seed}},
  title        = {Seed1.8: A generalized agentic model that can efficiently and accurately accomplish complex tasks in real-world scenarios},
  howpublished = {ByteDance Official Website},
  year         = {2025},
  url          = {https://seed.bytedance.com/en/seed1_8},
  note         = {Accessed: 2026-03-18}
}

@article{bai2025qwen3,
  title={Qwen3-vl technical report},
  author={Bai, Shuai and Cai, Yuxuan and Chen, Ruizhe and Chen, Keqin and Chen, Xionghui and Cheng, Zesen and Deng, Lianghao and Ding, Wei and Gao, Chang and Ge, Chunjiang and others},
  journal={arXiv preprint arXiv:2511.21631},
  year={2025}
}

@article{wang2025internvl35,
  title={Internvl3. 5: Advancing open-source multimodal models in versatility, reasoning, and efficiency},
  author={Wang, Weiyun and Gao, Zhangwei and Gu, Lixin and Pu, Hengjun and Cui, Long and Wei, Xingguang and Liu, Zhaoyang and Jing, Linglin and Ye, Shenglong and Shao, Jie and others},
  journal={arXiv preprint arXiv:2508.18265},
  year={2025}
}

@article{zhang2025videollama,
  title={Videollama 3: Frontier multimodal foundation models for image and video understanding},
  author={Zhang, Boqiang and Li, Kehan and Cheng, Zesen and Hu, Zhiqiang and Yuan, Yuqian and Chen, Guanzheng and Leng, Sicong and Jiang, Yuming and Zhang, Hang and Li, Xin and others},
  journal={arXiv preprint arXiv:2501.13106},
  year={2025}
}

@article{zhang2024llava,
  title={Llava-video: Video instruction tuning with synthetic data},
  author={Zhang, Yuanhan and Wu, Jinming and Li, Wei and Li, Bo and Ma, Zejun and Liu, Ziwei and Li, Chunyuan},
  journal={arXiv preprint arXiv:2410.02713},
  year={2024}
}

@article{xu2025qwen3,
  title={Qwen3-omni technical report},
  author={Xu, Jin and Guo, Zhifang and Hu, Hangrui and Chu, Yunfei and Wang, Xiong and He, Jinzheng and Wang, Yuxuan and Shi, Xian and He, Ting and Zhu, Xinfa and others},
  journal={arXiv preprint arXiv:2509.17765},
  year={2025}
}

@article{bain2023whisperx,
  title={Whisperx: Time-accurate speech transcription of long-form audio},
  author={Bain, Max and Huh, Jaesung and Han, Tengda and Zisserman, Andrew},
  journal={arXiv preprint arXiv:2303.00747},
  year={2023}
}

@article{shao2024visual,
  title={Visual cot: Advancing multi-modal language models with a comprehensive dataset and benchmark for chain-of-thought reasoning},
  author={Shao, Hao and Qian, Shengju and Xiao, Han and Song, Guanglu and Zong, Zhuofan and Wang, Letian and Liu, Yu and Li, Hongsheng},
  journal={Advances in Neural Information Processing Systems},
  volume={37},
  pages={8612--8642},
  year={2024}
}

@article{li2024benchmarking,
  title={Benchmarking multimodal retrieval augmented generation with dynamic vqa dataset and self-adaptive planning agent},
  author={Li, Yangning and Li, Yinghui and Wang, Xinyu and Jiang, Yong and Zhang, Zhen and Zheng, Xinran and Wang, Hui and Zheng, Hai-Tao and Yu, Philip S and Huang, Fei and others},
  journal={arXiv preprint arXiv:2411.02937},
  year={2024}
}

@article{rafailov2023direct,
  title={Direct preference optimization: Your language model is secretly a reward model},
  author={Rafailov, Rafael and Sharma, Archit and Mitchell, Eric and Manning, Christopher D and Ermon, Stefano and Finn, Chelsea},
  journal={Advances in neural information processing systems},
  volume={36},
  pages={53728--53741},
  year={2023}
}

@article{schulman2017proximal,
  title={Proximal policy optimization algorithms},
  author={Schulman, John and Wolski, Filip and Dhariwal, Prafulla and Radford, Alec and Klimov, Oleg},
  journal={arXiv preprint arXiv:1707.06347},
  year={2017}
}

@article{shao2024deepseekmath,
  title={Deepseekmath: Pushing the limits of mathematical reasoning in open language models},
  author={Shao, Zhihong and Wang, Peiyi and Zhu, Qihao and Xu, Runxin and Song, Junxiao and Bi, Xiao and Zhang, Haowei and Zhang, Mingchuan and Li, YK and Wu, Yang and others},
  journal={arXiv preprint arXiv:2402.03300},
  year={2024}
}

@article{abootorabi2025ask,
  title={Ask in any modality: A comprehensive survey on multimodal retrieval-augmented generation},
  author={Abootorabi, Mohammad Mahdi and Zobeiri, Amirhosein and Dehghani, Mahdi and Mohammadkhani, Mohammadali and Mohammadi, Bardia and Ghahroodi, Omid and Baghshah, Mahdieh Soleymani and Asgari, Ehsaneddin},
  journal={Findings of the Association for Computational Linguistics: ACL 2025},
  pages={16776--16809},
  year={2025}
}

@inproceedings{zhao2023retrieving,
  title={Retrieving multimodal information for augmented generation: A survey},
  author={Zhao, Ruochen and Chen, Hailin and Wang, Weishi and Jiao, Fangkai and Do, Xuan Long and Qin, Chengwei and Ding, Bosheng and Guo, Xiaobao and Li, Minzhi and Li, Xingxuan and others},
  booktitle={Findings of the Association for Computational Linguistics: EMNLP 2023},
  pages={4736--4756},
  year={2023}
}

\appendix

\newpage

\section{Data Construction Details}
\subsection{Video Collection}
\paragraph{\textbf{License}} All raw videos were collected from YouTube. To support non-commercial academic research and reduce copyright concerns, we restricted our collection to videos marked as distributed under the CC BY-NC 4.0 license.
\paragraph{\textbf{Ethical Considerations}} Our dataset is intended solely for non-commercial academic research on sports video understanding and officiating decision-making. Since the videos are drawn from real-world competitive matches, some clips may include strong physical contact, injury-related incidents, or emotionally intense confrontations. These contents are part of the natural context of professional sports and are retained to preserve the realism of officiating scenarios. At the same time, users of the dataset should handle such content responsibly and avoid uses that could promote harassment, athlete-targeted abuse, or misleading out-of-context reinterpretation of the original events.

\subsection{QA Annotation}
\paragraph{\textbf{Annotator Details}} To protect expert annotators from potential harassment or undue pressure associated with officiating judgments, all annotators remained anonymous throughout the annotation process. Annotators were compensated at a rate of \$12.50 per hour, which met or exceeded the relevant local minimum wage standard.

\paragraph{\textbf{Annotation Details}} Before formal annotation, all candidates were trained to use the annotation platform and were required to complete a pilot annotation task. Only candidates who passed this qualification stage were allowed to participate in the formal annotation process. For quality control and dispute resolution, the most experienced annotator—identified based on prior real-world refereeing experience and pilot-task performance—was designated as the senior referee to resolve disagreements.

\subsection{Text-Only Referee Exam Questions}
To further assess sports refereeing knowledge in a text-only setting, we collect a set of professional referee exam questions covering 11 sports. In total, the collected set contains 2,684 questions. The questions are gathered from publicly available online sources, including Wikipedia\footnote{\url{https://www.wikipedia.org/}}, Google\footnote{\url{https://www.google.com/}}, and Baidu\footnote{\url{https://www.baidu.com/}}. For non-English exam questions, we translate them into English using Gemini-3-Flash. Most questions come with official or publicly available reference answers. For the remaining items without such answers, we ask certified referees to provide answers, which are then used as the golden answers in our evaluation. 

Following international refereeing competency standards, we group the collected questions into three broad refereeing dimensions: \textbf{Rule Knowledge}, \textbf{Game Management}, and \textbf{Officiating Execution}. Table~\ref{tab:text_qa_distribution} summarizes the distribution of the collected text-only questions across sports and dimensions. The collected questions are not perfectly balanced across sports, mainly because referee exam materials are more accessible for some sports than for others. Most questions also fall into the Rule Knowledge category, with fewer in Game Management and Officiating Execution. This distribution is consistent with real-world referee examinations, which mainly emphasize rule mastery, making this setting reasonable for evaluating the rule-oriented knowledge of MLLMs.

\begin{table}[t]
    \centering
    \caption{Distribution of text-only referee exam questions across sports and categories.}
    \resizebox{\columnwidth}{!}{
    \begin{tabular}{lcccc}
        \toprule
        \textbf{Sport} & \textbf{Rule} & \textbf{Mgmt.} & \textbf{Exec.} & \textbf{OVERALL} \\
        \midrule
        Ice-Hockey & 208 & 16 & 46 & 270 \\
        Soccer & 408 & 11 & 22 & 441 \\
        Skating & 6 & 0 & 0 & 6 \\
        Tabletennis & 88 & 0 & 3 & 91 \\
        Handball & 232 & 14 & 8 & 254 \\
        Tennis & 371 & 22 & 22 & 415 \\
        Field-Hockey & 59 & 5 & 29 & 93 \\
        Volleyball & 333 & 18 & 46 & 397 \\
        Basketball & 474 & 8 & 14 & 496 \\
        Badminton & 121 & 5 & 17 & 143 \\
        Waterpolo & 72 & 2 & 4 & 78 \\
        \midrule
        \textbf{Total} & \textbf{2,372} & \textbf{101} & \textbf{211} & \textbf{2,684} \\
        \bottomrule
    \end{tabular}
    }
    \label{tab:text_qa_distribution}
\end{table}

\section{Detailed Experimental Settings}

\subsection{Hyperparameters}
Unless otherwise specified, we adopt a unified decoding configuration for all models during inference. Specifically, the temperature is set to 0.1, and top-\(p\) is set to 0.95.



\begin{table*}[htbp]
    \centering
    \caption{Gemini-3-Flash performance of different prompting strategies on RefereeBench. Results are reported by question category.}
    \resizebox{\textwidth}{!}{
    \begin{tabular}{l c ccc ccc c c c}
        \toprule
        \multirow{2}{*}{\textbf{Methods}} & \textbf{Existence (\%)} & \multicolumn{3}{c}{\textbf{Classification (\%)}} & \multicolumn{3}{c}{\textbf{Reasoning (\%)}} & \textbf{Perception (\%)} & \textbf{Grounding (\%)} & \multirow{2}{*}{\textbf{OVERALL (\%)}} \\
        \cmidrule(lr){2-2} \cmidrule(lr){3-5} \cmidrule(lr){6-8} \cmidrule(lr){9-9} \cmidrule(lr){10-10}
        & \textbf{Q1} & \textbf{Q2} & \textbf{Q4} & \textbf{Avg.} & \textbf{Q3} & \textbf{Q5} & \textbf{Avg.} & \textbf{Q6} & \textbf{Q7} & \\
        \midrule
        \textbf{Zero-Shot} & 71.2 & 53.2 & 49.6 & 51.4 & \textbf{48.1} & 52.5 & \textbf{50.3} & \textbf{79.6} & 42.8 & 56.7 \\
        \textbf{Role Prompting} & 70.8 & 51.9 & 41.2 & 46.6 & 46.5 & 52.1 & 49.3 & 77.4 & \textbf{48.6} & 55.5 \\
        \textbf{Generic Prompting} & \textbf{71.9} & 51.3 & 49.3 & 50.3 & 46.5 & 51.4 & 49.0 & 77.7 & 42.3 & 55.8 \\
        \textbf{Targeted Prompting} & 66.8 & \textbf{57.2} & \textbf{54.3} & \textbf{55.8} & 45.5 & \textbf{54.6} & 50.1 & 78.3 & 46.4 & \textbf{57.9} \\
        \bottomrule
    \end{tabular}
    }
    \label{tab:question}
\end{table*}

\begin{table*}[htbp]
    \centering
    \caption{Gemini-3-Flash performance of different prompting strategies on RefereeBench. Results are reported by six sport types.}
    \resizebox{\textwidth}{!}{
    \begin{tabular}{l c c c c c c c}
        \toprule
        \multirow{2}{*}{\textbf{Methods}} & \multicolumn{6}{c}{\textbf{Sports (\%)}} & \multirow{2}{*}{\textbf{OVERALL (\%)}} \\
        \cmidrule(lr){2-7}
        & \textbf{Basketball} & \textbf{Soccer} & \textbf{Tennis} & \textbf{Volleyball} & \textbf{Waterpolo} & \textbf{Handball} & \\
        \midrule
        \textbf{Zero-Shot} & 51.6 & \textbf{77.8} & 56.7 & 60.2 & \textbf{36.9} & 61.1 & 56.7 \\
        \textbf{Role Prompting} & \textbf{53.8} & 73.1 & 59.2 & 58.8 & 29.8 & 61.3 & 55.5 \\
        \textbf{Generic Prompting} & 53.1 & 75.4 & 59.2 & \textbf{61.3} & 32.1 & 58.4 & 55.8 \\
        \textbf{Targeted Prompting} & 53.7 & 77.2 & \textbf{59.7} & 60.3 & 34.3 & \textbf{65.1} & \textbf{57.9} \\
        \bottomrule
    \end{tabular}
    }
    \label{tab:sports}
\end{table*}

\subsection{Frame Extraction}
During inference, we choose to convert each video into a sequence of sampled frames and provide the resulting multi-image inputs to the model. For models that naturally support frame-rate-based sampling, we sample frames at 1 frame per second. For all other models, we adopt fixed-size uniform sampling according to their practical inference settings. The specific numbers of sampled frames are as follows: 1 frame per second for Gemini 3 Flash, Gemini 3 Pro, and Qwen3-VL-8B; 50 frames for GPT-4o and GPT-5; 100 frames for Claude-4.5-Haiku and Claude-4.5-Sonnet; 92 frames for Doubao-Seed-1.8; 64 frames for InternVL3.5-8B; 32 frames for LLaVA-Video-7B; and 180 frames for VideoLLaMA3-7B.

All videos are evaluated at 720p resolution. Since the original videos are either 720p or 1080p, only the 1080p videos are downsampled, while the 720p videos remain unchanged. For omni models, we use the same resolution setting but directly encode the full video clips during inference, without manual frame extraction.

\subsection{Evaluation Prompts}
For all models, we use a unified system prompt:
\begin{quote}
You are a helpful assistant. Your task is to provide an accurate answer and a concise explanation for the given objective questions based on the video.
\end{quote}

and the user prompt template:
\begin{quote}
Select the best answer to the following multiple choice question based on the video. Respond with only the letter (A, B, C, or D) of the correct option. [Question]: \{question\} [Options]: \{options\} The best answer is:
\end{quote}

\begin{table*}[htbp]
    \centering
    \caption{Gemini-3-Flash performance of rule-augmented inference across 11 sports. We compare the Zero-Shot baseline with two retrieval-augmented settings, namely Plain-RAG and Hybrid-RAG, on rule-intensive question types (Q2--Q5).}
    \resizebox{\textwidth}{!}{
    \begin{tabular}{l ccccc c ccccc c ccccc}
        \toprule
        \multirow{2}{*}{\textbf{Sports}} & \multicolumn{5}{c}{\textbf{Zero-Shot}} & & \multicolumn{5}{c}{\textbf{Plain-RAG}} & & \multicolumn{5}{c}{\textbf{Hybrid-RAG}} \\
        \cmidrule(lr){2-6} \cmidrule(lr){8-12} \cmidrule(lr){14-18}
        & \textbf{Q2} & \textbf{Q3} & \textbf{Q4} & \textbf{Q5} & \textbf{Overall} & & \textbf{Q2} & \textbf{Q3} & \textbf{Q4} & \textbf{Q5} & \textbf{Overall} & & \textbf{Q2} & \textbf{Q3} & \textbf{Q4} & \textbf{Q5} & \textbf{Overall} \\
        \midrule
        Ice-Hockey & 78.79 & 63.64 & 90.91 & 74.75 & \textbf{77.02} & & 74.75 & 60.61 & 94.90 & 75.76 & 76.46 & & 54.00 & 50.00 & 68.00 & 76.00 & 62.00 \\
        Soccer & 89.87 & 82.28 & 78.48 & 68.35 & \textbf{79.75} & & 82.28 & 83.54 & 73.42 & 68.35 & 76.90 & & 76.00 & 80.00 & 62.00 & 66.00 & 71.00 \\
        Skating & 62.50 & 39.29 & 78.57 & 67.86 & \textbf{62.05} & & 67.86 & 33.93 & 75.00 & 60.71 & 59.38 & & 59.09 & 20.45 & 65.91 & 47.73 & 48.30 \\
        Tabletennis & 84.69 & 77.55 & 72.45 & 53.06 & 71.94 & & 77.55 & 75.51 & 73.47 & 42.86 & 67.35 & & 78.00 & 82.00 & 73.47 & 56.00 & \textbf{72.36} \\
        Handball & 59.57 & 37.23 & 45.74 & 51.06 & \textbf{48.40} & & 52.13 & 28.72 & 34.04 & 46.81 & 40.43 & & 38.00 & 32.00 & 36.00 & 42.86 & 37.19 \\
        Tennis & 67.78 & 63.33 & 37.78 & 63.33 & 58.06 & & 63.33 & 73.33 & 36.67 & 68.89 & \textbf{60.56} & & 70.00 & 78.00 & 32.00 & 64.00 & 61.00 \\
        Field-Hockey & 34.29 & 31.43 & 74.29 & 43.81 & \textbf{45.95} & & 34.29 & 35.24 & 74.29 & 25.71 & 42.38 & & 34.00 & 30.00 & 74.00 & 36.00 & 43.50 \\
        Volleyball & 63.16 & 52.63 & 65.79 & 69.74 & 62.83 & & 73.68 & 42.11 & 73.68 & 75.00 & \textbf{66.12} & & 48.00 & 53.06 & 22.00 & 61.22 & 45.96 \\
        Basketball & 29.09 & 31.82 & 50.00 & 36.11 & 36.76 & & 29.09 & 38.18 & 41.82 & 37.04 & 36.53 & & 32.00 & 42.00 & 34.00 & 46.00 & \textbf{38.50} \\
        Badminton & 37.93 & 48.28 & 27.59 & 75.86 & 47.41 & & 65.52 & 51.72 & 24.14 & 93.10 & \textbf{58.62} & & 41.38 & 48.28 & 20.69 & 82.76 & 48.28 \\
        Waterpolo & 20.22 & 30.34 & 25.84 & 34.09 & 27.61 & & 23.60 & 13.48 & 16.85 & 29.55 & 20.85 & & 36.00 & 30.00 & 18.00 & 40.82 & \textbf{31.16} \\
        \midrule
        \textbf{Average (\%)} & \textbf{57.19} & \textbf{50.49} & \textbf{60.32} & \textbf{55.64} & \textbf{55.91} & & 56.54 & 48.65 & 57.58 & 52.93 & 53.92 & & 51.82 & 50.00 & 46.74 & 55.38 & 50.98 \\
        \bottomrule
    \end{tabular}
    }
    \label{tab:rag_impact}
\end{table*}

\section{Additional Analysis}

We further investigate whether stronger prompting strategies and external rule augmentation can improve model performance on automatic sports refereeing. Overall, both analyses suggest that these interventions bring only limited gains and do not fundamentally resolve the core challenges posed by RefereeBench.

\subsection{Prompting Strategy Analysis}

\paragraph{\textbf{Experimental Setting.}} We study whether prompt design can improve model performance on automatic sports refereeing. We first select six representative sports, including Basketball, Soccer, Tennis, Volleyball, Waterpolo, and Handball, for a controlled comparison of prompting strategies, and then evaluate the selected strategy on Gemini-3-Flash over the full RefereeBench benchmark. To this end, we compare four prompting strategies: \textbf{Zero-Shot}, \textbf{Role Prompting}, \textbf{Generic Prompting}, and \textbf{Targeted Prompting}. Zero-Shot directly asks the model to answer the question based on the video. Role Prompting further assigns the model the identity of an experienced international referee, with the goal of introducing domain-specific judgment into the decision process. Generic Prompting uses a unified four-step reasoning chain that encourages the model to perceive the scene, recall the rule, locate the evidence, and then make the decision. In contrast, Targeted Prompting is tailored to the seven question types in RefereeBench, so that different categories of questions are matched with different reasoning procedures. Specifically, existence and perception questions emphasize direct visual identification, classification questions focus on matching the observed interaction to the closest candidate label, reasoning questions require identifying the key event before selecting the best explanation, and temporal grounding questions focus on locating the decisive moment.

\paragraph{\textbf{Results and Discussion.}} The results in Table\ref{tab:question} and Table\ref{tab:sports} show that prompt design has a measurable but limited effect on sports officiating performance. In the six-sport comparison, Targeted Prompting (57.9\%) achieves the best overall accuracy, outperforming Zero-Shot (56.7\%), Generic Prompting (55.8\%), and Role Prompting (55.5\%). Based on this result, we adopt Targeted Prompting in the subsequent analysis. The detailed breakdown further shows that the main gains come from classification-related questions. Compared with Zero-Shot, Targeted Prompting improves Q2 from 53.2\% to 57.2\% and Q4 from 49.6\% to 54.3\%, raising the classification average from 51.4\% to 55.8\%. By contrast, the changes on existence, reasoning, perception, and grounding are relatively small, and some categories show slight declines. Overall, these results suggest that fine-grained prompt design is most helpful for tasks that require precise matching between visual evidence and predefined officiating categories, but its overall improvement remains modest.

\subsection{Analysis of Rule-Augmented Inference}

\paragraph{\textbf{Experimental Setting.}} We next investigate whether external rule retrieval can improve performance on sports officiating tasks. Since rule augmentation is most relevant to decisions that require explicit rule grounding, we focus on rule-intensive question types, including Q2--Q5. We evaluate \textit{Gemini-3-Flash-Preview} under three settings: \textbf{Zero-Shot}, \textbf{Plain-RAG}, and \textbf{Hybrid-RAG}. In the Zero-Shot setting, the model answers questions using only the video and its internal parametric knowledge. Plain-RAG is our domain-specific retrieval pipeline designed to reduce interference between noisy visual descriptions and legal rule text. Specifically, we first collect official rulebooks for 11 sports, parse them into rule chunks, and index them in a sport-specific vector database. During retrieval, we deliberately exclude video descriptions from the initial recall stage and instead construct multiple text queries from the question and candidate options, so that retrieval is driven primarily by rule semantics rather than potentially noisy visual captions. For sanction-related questions, we further introduce sport-specific textual anchors to better capture relevant penalty clauses. After candidate rule chunks are retrieved and deduplicated, the visual modality is introduced only in the reranking stage, where each candidate chunk is scored against queries constructed from the video observation, question, and answer options. This asymmetric design keeps retrieval text-centered in the early stage while still incorporating visual evidence during final ranking. In contrast, Hybrid-RAG is a general-purpose RAG baseline that combines structured parsing, hybrid retrieval, and reranking; in implementation, we use the LlamaIndex framework.\footnote{\url{https://github.com/run-llama/llama_index}}

\paragraph{\textbf{Results and Discussion.}} The results show in Table\ref{tab:rag_impact} that external rule retrieval does not consistently improve end-to-end sports officiating performance. Averaged across 11 sports, Zero-Shot (55.91\%) achieves the best overall accuracy, while Plain-RAG (53.92\%) performs worse and Hybrid-RAG (50.98\%) drops further. This trend is especially clear on classification-heavy questions. For example, on Q2, the average accuracy decreases from Zero-Shot (57.19\%) to Plain-RAG (56.54\%) and Hybrid-RAG (51.82\%); on Q4, it declines from Zero-Shot (60.32\%) to Plain-RAG (57.58\%) and Hybrid-RAG (46.74\%). These results suggest that simply introducing retrieved rule text does not reliably help visually grounded officiating decisions. A closer analysis indicates that the main difficulty lies in how retrieved legal knowledge interacts with visual evidence during final decision making. When the model fails to perceive the key physical event correctly, even the correct rule cannot produce the correct answer, because it is applied to an incorrect visual premise. When the visual evidence is ambiguous, retrieved rule text may further bias the model toward over-interpreting the scene and forcing the observation to fit the injected legal criteria. Overall, these findings suggest that naive rule augmentation is insufficient for this benchmark, and that more effective approaches will likely require more selective and more tightly visually grounded rule integration.

\section{Case Study}

\begin{figure*}
  \includegraphics[width=\textwidth]{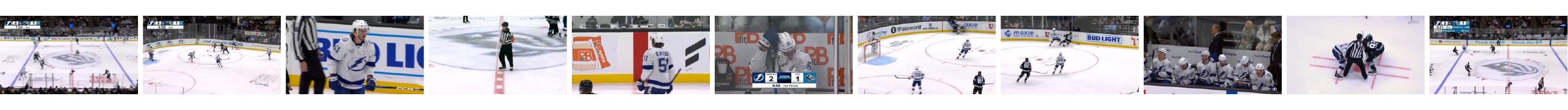}
  \caption{Representative instance from the Ice-Hockey subset. The foul shown is \textit{Tripping} and the penalty is \textit{Minor Penalty}.}
  \label{fig:case_study_icehockey}
\end{figure*}

\begin{figure*}
  \includegraphics[width=\textwidth]{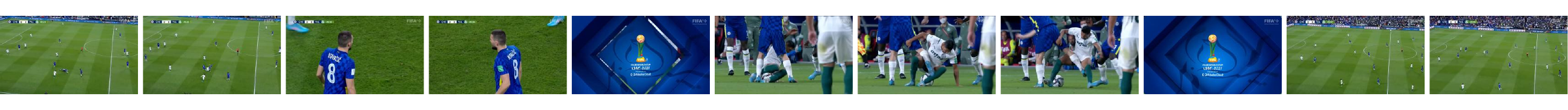}
  \caption{Representative instance from the Soccer subset. The foul shown is \textit{Unsporting behaviour} and the penalty is \textit{Direct free kick}.}
  \label{fig:case_study_soccer}
\end{figure*}

\begin{figure*}
  \includegraphics[width=\textwidth]{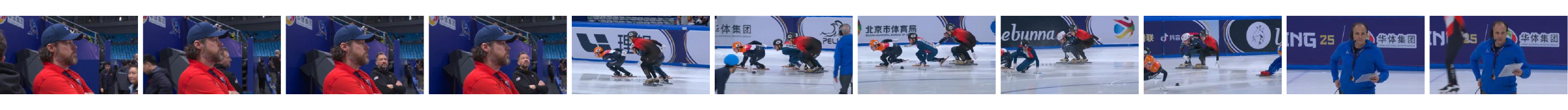}
  \caption{Representative instance from the Skating subset. The foul shown is \textit{Lane Change from Outside to In} and the penalty is \textit{Penalty Advancement}.}
  \label{fig:case_study_skating}
\end{figure*}

\begin{figure*}
  \includegraphics[width=\textwidth]{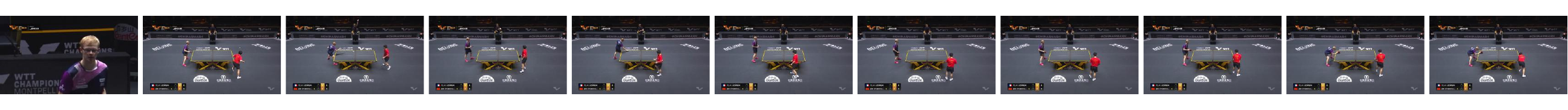}
  \caption{Representative instance from the Tabletennis subset. The foul shown is \textit{Service Foul} and the penalty is \textit{Warning}.}
  \label{fig:case_study_tabletennis}
\end{figure*}

\begin{figure*}
  \includegraphics[width=\textwidth]{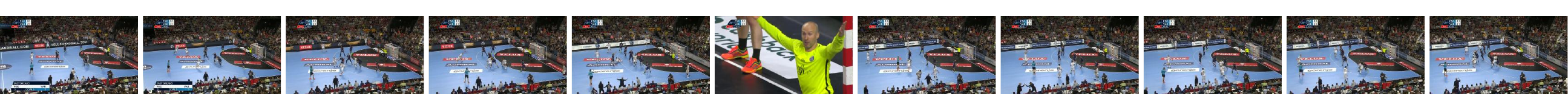}
  \caption{Representative instance from the Handball subset. The foul shown is \textit{Pushing} and the penalty is \textit{Free Throw}.}
  \label{fig:case_study_handball}
\end{figure*}

\begin{figure*}
  \includegraphics[width=\textwidth]{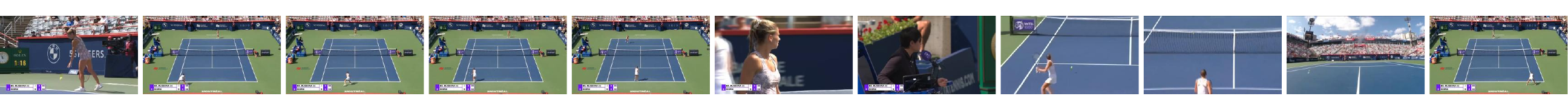}
  \caption{Representative instance from the Tennis subset. The foul shown is \textit{Double Bounce} and the penalty is \textit{Point Awarded To Opponent}.}
  \label{fig:case_study_tennis}
\end{figure*}

\begin{figure*}
  \includegraphics[width=\textwidth]{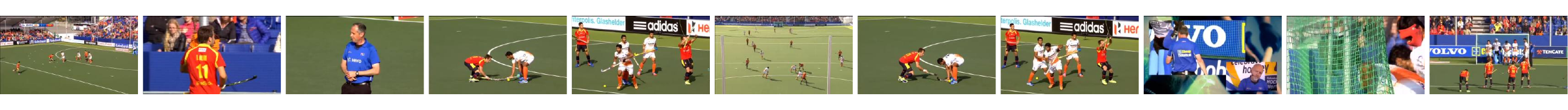}
  \caption{Representative instance from the Field-Hockey subset. The foul shown is \textit{Intentional foul} and the penalty is \textit{Penalty Corner Free Hit}.}
  \label{fig:case_study_fieldhockey}
\end{figure*}

\begin{figure*}
  \includegraphics[width=\textwidth]{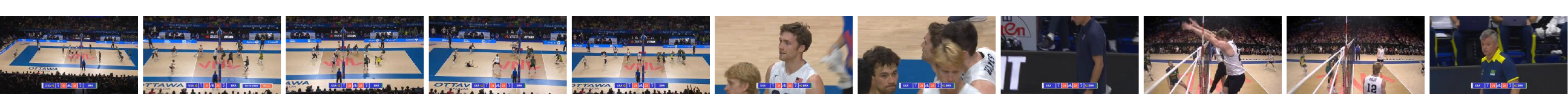}
  \caption{Representative instance from the Volleyball subset. The foul shown is \textit{Net Touch} and the penalty is \textit{Point Awarded To Opponent}.}
  \label{fig:case_study_volleyball}
\end{figure*}

\begin{figure*}
  \includegraphics[width=\textwidth]{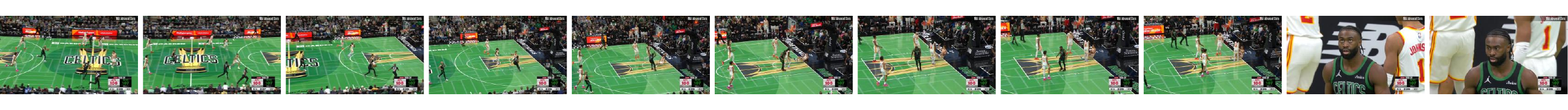}
  \caption{Representative instance from the Basketball subset. The foul shown is \textit{Blocking} and the penalty is \textit{Free Throw (2)}.}
  \label{fig:case_study_basketball}
\end{figure*}

\begin{figure*}
  \includegraphics[width=\textwidth]{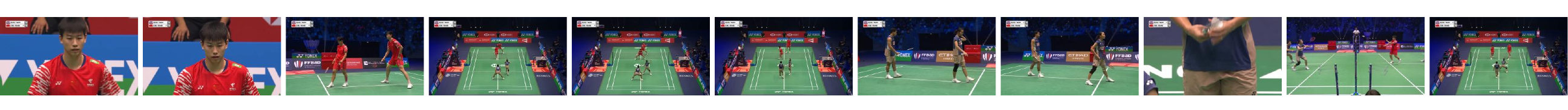}
  \caption{Representative instance from the Badminton subset. The foul shown is \textit{Service Fault} and the penalty is \textit{Point Awarded To Opponent}.}
  \label{fig:case_study_badminton}
\end{figure*}

\begin{figure*}
  \includegraphics[width=\textwidth]{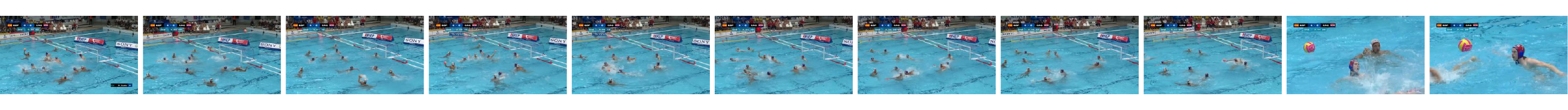}
  \caption{Representative instance from the Waterpolo subset. The foul shown is \textit{Ordinary Foul} and the penalty is \textit{Free Throw}.}
  \label{fig:case_study_waterpolo}
\end{figure*}

To provide a more concrete illustration of RefereeBench, we present qualitative examples spanning all 11 sports in our dataset. These case studies highlight both the visual diversity and the domain-specific reasoning challenges inherent to the benchmark. The corresponding examples are detailed in Figures~\ref{fig:case_study_icehockey} through~\ref{fig:case_study_waterpolo}.

\end{document}